\documentclass[runningheads]{llncs}

\usepackage{eccv}

\newcommand{\RNum}[1]{\uppercase\expandafter{\romannumeral #1\relax}}

\usepackage{eccvabbrv}

\usepackage{graphicx}
\usepackage{booktabs}
\usepackage{makecell}
\usepackage{amsmath}
\usepackage{bibunits}

\usepackage[accsupp]{axessibility}  

\usepackage[breaklinks,colorlinks]{hyperref}
\usepackage{multirow}

\usepackage{orcidlink}

\defaultbibliography{main}
\defaultbibliographystyle{splncs04}

\usepackage{setspace}

\let\titleold\title
\let\instituteold\institute
\let\authorold\author

\renewcommand{\title}[1]{\titleold{#1}\newcommand{\thetitle}{#1}}
\renewcommand{\institute}[1]{\instituteold{#1}\newcommand{\theinstitute}{#1}}
\renewcommand{\author}[1]{\authorold{#1}\newcommand{\theauthor}{#1}}

\newcommand{\maketitlesupplementaryfull}
   {
    \subtitle{Supplementary Material\vspace{-1em}}
    \titleold{\thetitle}
    \authorold{\theauthor}
    \instituteold{\theinstitute}
    \maketitle
   }

\begin{document}

\title{MetaWeather: Few-Shot Weather-Degraded Image Restoration} 


\author{Youngrae Kim$^\star$
\and
Younggeol Cho\thanks{These authors contributed equally.}
\and
Thanh-Tung Nguyen
\and
Seunghoon Hong 
\and
Dongman Lee
}

\authorrunning{Y. Kim et al.}%

\institute{School of Computing, Korea Advanced Institute of Science and Technology (KAIST)
\email{\{youngrae.kim, rangewing, tungnt, seunghoon.hong, dlee\}@kaist.ac.kr}}

%
\maketitle

\begin{bibunit}

\begin{abstract}
Real-world weather conditions are intricate and often occur concurrently. 
However, most existing restoration approaches are limited in their applicability to specific weather conditions in training data and struggle to generalize to unseen weather types, including real-world weather conditions.
To address this issue, we introduce MetaWeather, a universal approach that can handle diverse and novel weather conditions with a single unified model.
Extending a powerful meta-learning framework, MetaWeather formulates the task of weather-degraded image restoration as a few-shot adaptation problem that predicts the degradation pattern of a query image, and learns to adapt to unseen weather conditions through a novel spatial-channel matching algorithm. 
Experimental results on the BID Task II.A, SPA-Data, and RealSnow datasets demonstrate that the proposed method can adapt to unseen weather conditions, significantly outperforming the state-of-the-art multi-weather image restoration methods.

\end{abstract}
\section{Introduction}
\label{sec:intro}

\par
Adverse weather conditions like rain, snow, and fog obfuscate visual clarity and degrade the quality of captured images and footage. Such degradation significantly and directly affects the performance of diverse vision tasks, including object detection, tracking, segmentation, and depth estimation~\cite{Huang2020DSNetWeatherOD,Liu2022YOLOweather,Hassaballah2021BadWeatherTracking,Tremblay2021BadWeatherSegmentDepth}. The impact is especially critical for real-time outdoor vision applications, such as autonomous driving and surveillance, causing the systems to yield unreliable results.
Therefore, weather-degraded image restoration has attracted much attention from the research community in recent years. Recently, several works \cite{li2020all,valanarasu2022transweather,zhu2023learning, ye2023adverse} have proposed methods for restoring images degraded by multiple types of weather conditions.

\par
\begin{figure}[t]
    \centering
    \begin{subfigure}{0.49\linewidth}
        \centering
        \includegraphics[width=\linewidth]{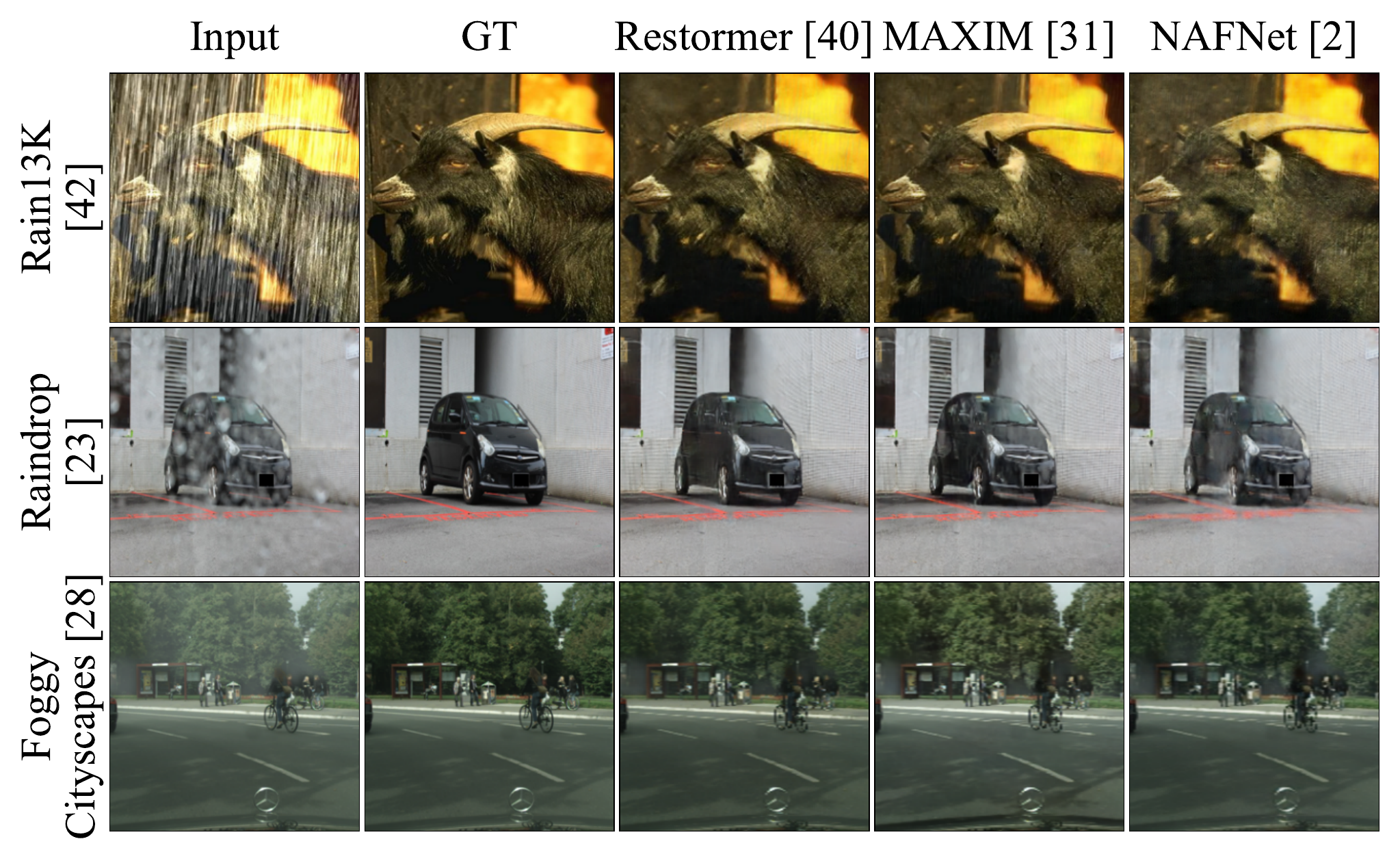}
        \label{fig:fig1a}
        \caption{Seen weather conditions}
    \end{subfigure}
    \hfill
    \begin{subfigure}{0.49\linewidth}
        \centering
        \includegraphics[width=\linewidth]{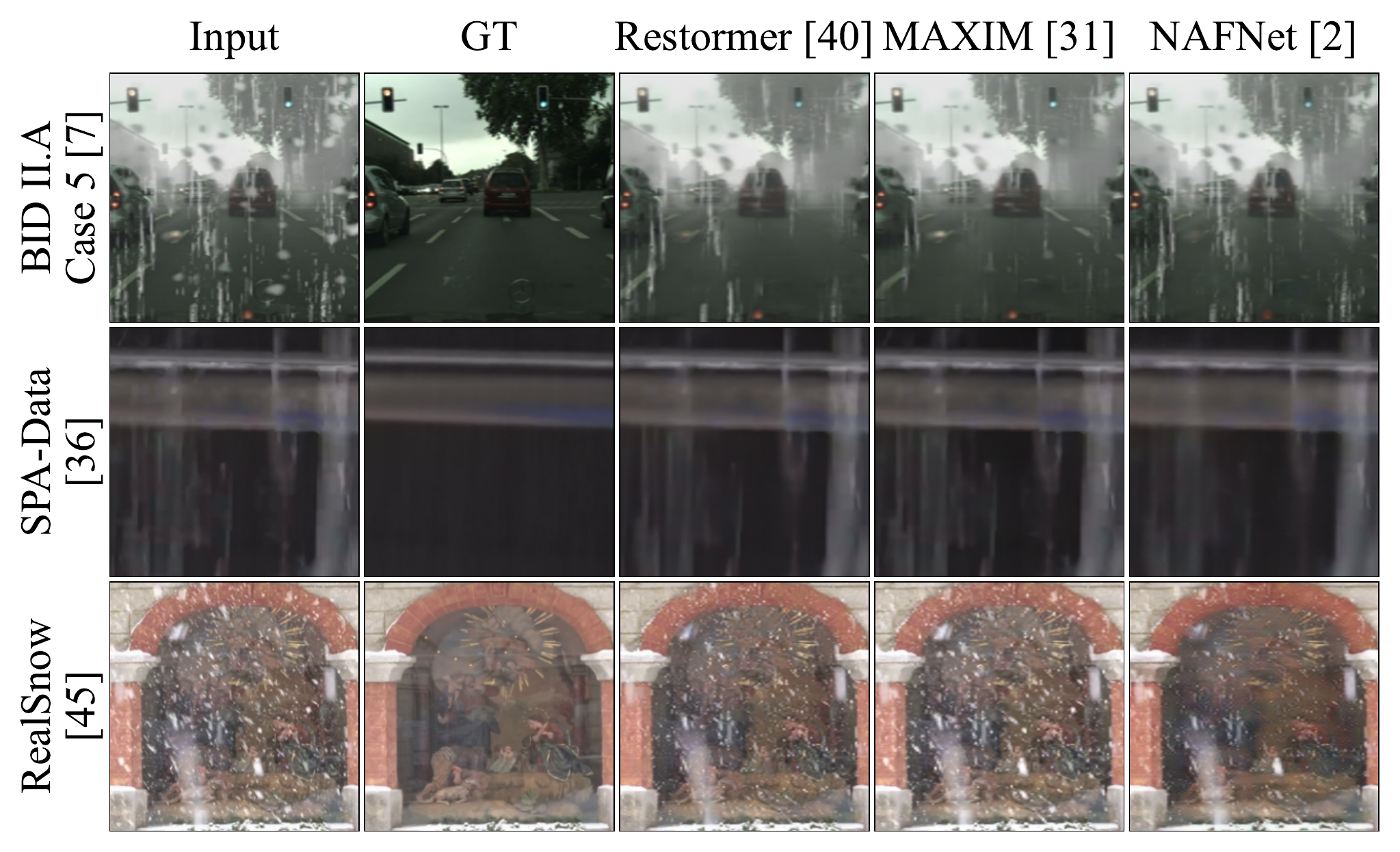}
        \label{fig:fig1b}
        \caption{Unseen weather conditions}
    \end{subfigure}
    \caption{
    (a). The state-of-the-art image restoration models perform well on the seen weather types in the training data, including rain~\cite{zamir2021multi}, snow~\cite{Liu2018desnow}, raindrop~\cite{qian2018attentive}, and fog~\cite{sakaridis2018semantic}. (b). In contrast, these models show significant performance degradation on unseen weather types outside the scope of the training data, such as real rain (SPA-Data~\cite{Wang2019spa}) and real snow (RealSnow~\cite{zhu2023learning}), as well as co-occurrence of the weather types, rain, fog, and raindrops, in the training data (BID \RNum{2}.A Case 5~\cite{han2022blind}). Best viewed with zoom and color.
    }
    \label{fig:fig1}
\end{figure}

However, existing multi-weather degraded image restoration methods are unable to handle unseen weather conditions beyond the scope of their training data. 
\cref{fig:fig1} shows that multi-weather degraded image restoration methods jointly trained on single-condition synthetic datasets, including rain~\cite{zamir2021multi}, fog~\cite{sakaridis2018semantic}, snow~\cite{Liu2018desnow}, and raindrops~\cite{qian2018attentive}, cannot deal with the unseen weather phenomena, which includes the co-occurrence of the trained conditions and the real weather-degraded footage.
Our findings indicate that existing approaches do not adequately handle unseen weather conditions.
A domain translation-based method \cite{patil2023multi} has been presented to cope with the real-world weather conditions. However, the method is unable to handle all cases of unseen weather conditions, as it only addresses the domain gap between synthetic and real data, rather than addressing more complex weather conditions.
Restoring images degraded by unseen and arbitrary weather conditions has been under-explored, despite the imperative need.

To effectively restore such weather-degraded images, the following considerations should be met.
First, the method should be data-efficient. In the real world, obtaining a sufficient number of clean images is often impractical. Several studies~\cite{Wang2019spa, ba2022gtrain} have proposed filter-based methods for acquiring weather-degraded image pairs from real-world videos. However, due to the use of a stationary camera, they can only obtain one or a few pairs of degraded and clean images from a single location.
Second, the method should be weather-agnostic. This is crucial because the deployed image restoration model may encounter various types of weather in real-world scenarios, including those outside the scope of the training data.




%


\par
In this paper, we propose \textbf{MetaWeather}, a novel method for restoring images degraded by any unseen weather conditions with a single unified model. 
We treat the problem of image restoration for unseen weather as a few-shot adaptation problem, fully exploiting a small number of support pairs.
MetaWeather extends a flexible and universal meta-learning framework for dense prediction~\cite{kim2023universal} to few-shot weather degraded image restoration.
Specifically, given a few-shot support set of clean and degraded images, MetaWeather extracts the common degradation patterns of a query image, such as rain streaks and fog patterns, through spatial-channel matching.
Then the identified degradation patterns are used to restore the query image.
Our experiments demonstrate that matching the degradation patterns guides the architecture to learn more representation over a few-shot support set, thereby enabling more flexible adaptation. Additionally, we also observe that the spatial-channel matching better extracts the degradation patterns, where the spatial and channel dimensions primarily deal with weather particles and atmospheric effects, respectively.

\par
Our contributions are summarized as follows:
\begin{itemize}
    \item We propose MetaWeather, a novel few-shot weather-degraded image restoration method that enables restoration under any unseen and weather conditions. To the best of our knowledge, this is the first work introducing few-shot learning to the image restoration task for unseen weather conditions without any prior assumption.
    \item We extend the matching-based meta-learning framework to extract degradation patterns through a novel matching algorithm, namely spatial-channel matching, which effectively leverages weather-related representations from a few-shot support set. 
    \item We evaluate MetaWeather on the BID Task \RNum{2}.A \cite{han2022blind}, SPA-Data~\cite{Wang2019spa}, and RealSnow \cite{zhu2023learning} datasets containing weather-degraded images under various combined weather conditions, real rain, and real snow, respectively. Extensive experimental results demonstrate that the proposed method effectively adapts to any unseen weather conditions, outperforming the state-of-the-art multi-weather image restoration methods.
\end{itemize}

\section{Related Works}
\label{sec:related}

\subsection{Generic Image Restoration}

To handle multiple weather conditions in a single model, All-in-One~\cite{li2020all} first introduces a generic model that has the ability to handle multiple weather degradations by using an individual encoder for each, which leads to high computational complexity.
To eliminate the limitation of All-in-One, different methods~\cite{ye2023adverse,valanarasu2022transweather,zhu2023learning,ozdenizci2023weatherdiffusion} are proposed, which can handle several weather degradation patterns with a generic encoder and decoder. 
Similarly, generic models have been proposed to handle various restoration tasks, such as denoising, deblurring, and enhancement~\cite{zamir2022restormer,chen2022simple,tu2022maxim,li2022all,zhao2023comprehensive,zhou2023fourmer,cui2023focal}. 
As they possess versatile architectures capable of handling a multitude of degradation types, these networks have demonstrated outstanding performances across diverse restoration tasks.
However, despite their efficacy, these models exhibit notable performance deterioration when encountering new patterns of degradation as shown in \cref{fig:fig1}.


\subsection{Few-shot Learning on Image Restoration}
Recently, few-shot learning approaches are exploited in various vision tasks for mitigating the label scarcity problem~\cite{vinyals2016matching,kang2019few,ran2023few}. 
Within these, the ones closest to image restoration are dense prediction tasks ~\cite{kim2023universal}, such as semantic segmentation~\cite{shaban2017one,wang2019panet}.
Although few-shot learning methods have been widely explored in such tasks, it is challenging to directly apply the proposed methods to image restoration tasks. This is because the image restoration model needs to learn how to preserve fine spatial details to restore the image \cite{zamir2020learning}, and it is a challenging to generalize these details with only a few images. 

Prior efforts to incorporate few-shot learning into image restoration have been presented. MLDN~\cite{gao2021meta} employ a meta-learning methodology to establish connections between rainy and clean images, enabling adaptation to new rain patterns. Similarly, FLUID~\cite{rai2022fluid} introduces a framework for few-shot adaptation in the deraining task. Although FLUID exhibits adaptability in rainy conditions, its focus is limited to rainy weather scenarios where a prior degradation distribution is presumed. Similarly, Liu \textit{et al.}~\cite{Liu2022testtimedehaze} propose a model agnostic meta-learning~\cite{finn2017model} based method for the dehazing task, enabling adaptable responses to new types of haze. These works demonstrate successful attempts to introduce few-shot learning in weather-degraded image restoration tasks. However, it is important to note that these methods presuppose knowledge of future degradation types, allowing them to establish predefined knowledge aimed at expected weather conditions. In contrast, our proposed method maximizes the utilization of degradation patterns within few-shot images, facilitating the flexible transfer of model representation to new degradation patterns, and thus avoiding task type limitations or requirements for prior degradation information.





\section{MetaWeather}
\label{sec:method}


\subsection{Problem Formulation}
\label{subsec:problem}
Following the prior works~\cite{Liu2018desnow, Qian2018raindrop, li2019heavy}, we consider a degradation pattern that produces the degraded image $\mathbf{X} \in \mathbb{R}^{H \times W \times 3}$ from the clean image $\mathbf{Y} \in \mathbb{R}^{H \times W \times 3}$:
\begin{equation}
\label{eq:weather-degraded_image}
\mathbf{X}=\mathbf{T} \odot (\mathbf{Y} + \mathbf{P}) + (1-\mathbf{T})\odot \mathbf{A},
\end{equation}
where $\mathbf{T}$ is the transmission map, $\mathbf{P}$ is the sum of degradation particles, and $\mathbf{A}$ is the global atmospheric light of the scene. 



Our objective is to learn a function $\mathcal{F}: \mathbb{R}^{H\times W \times 3}\rightarrow \mathbb{R}^{H\times W \times 3}$ that produces a clean image $\mathbf{\hat{Y}}$ of a query image $\mathbf{X}$ degraded by an arbitrary weather condition $\omega$, given a few-shot {support} set of $N$ labeled pairs sampled from the same degradation patterns $\mathbf{S_\omega}=\{(\mathbf{X}_i, \mathbf{Y}_i)\}_{i\le N}$:

\begin{equation}
\label{eq:objective}
\mathbf{\hat{Y}}=\mathcal{F}(\mathbf{X};\mathbf{S_\omega}) .
\end{equation}

%



\begin{figure*}[t]
    \centering
    \includegraphics[width=\linewidth]{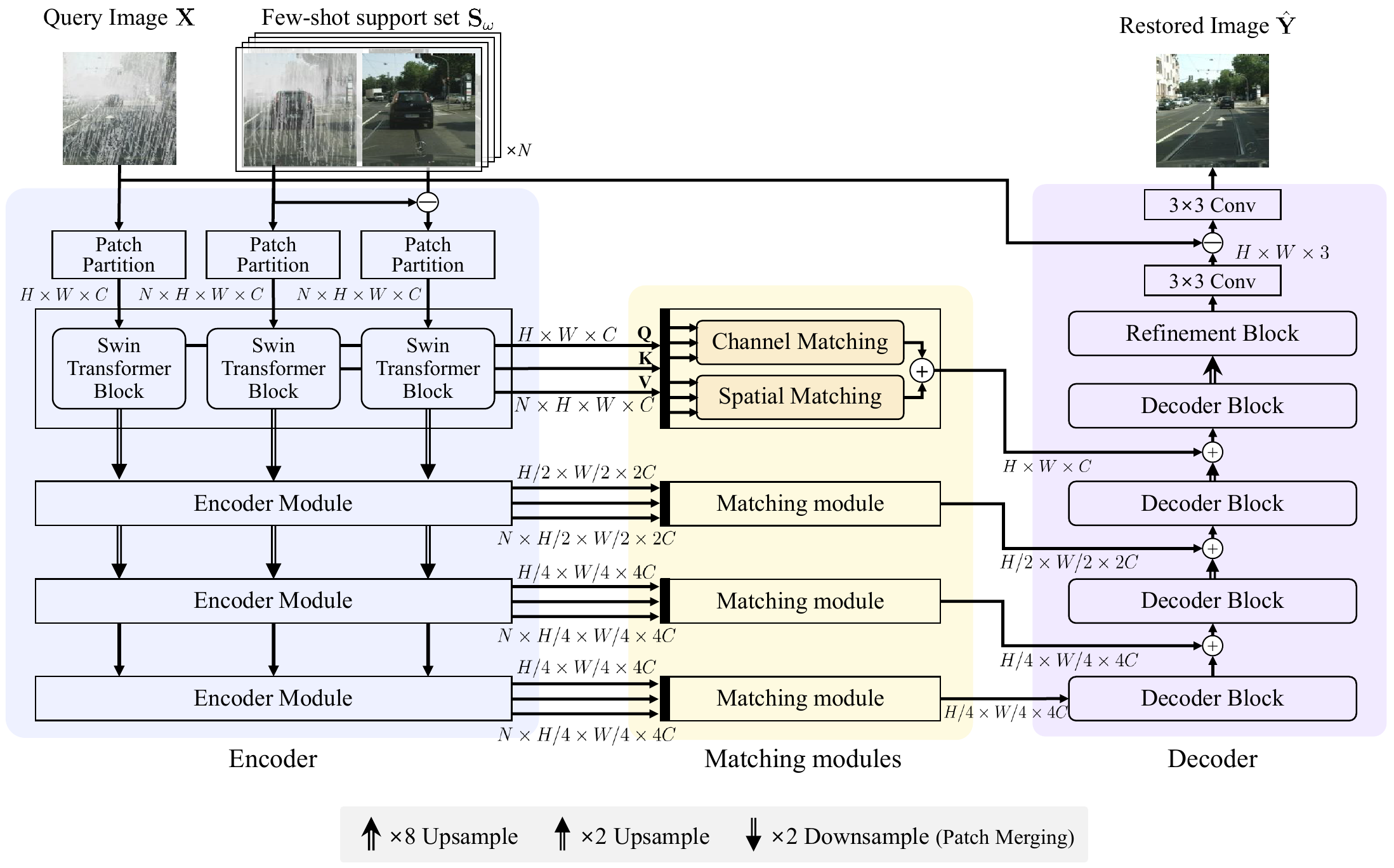}
    \caption{Overall architecture of MetaWeather. MetaWeather consists of a hierarchical encoder-decoder design and matching module. Our matching module matches the degradation pattern between query and support set images, which enables MetaWeather to fully utilize a few-shot support set. The matching results are passed on to the decoder blocks at each level, and the extracted degradation pattern of the query image is then subtracted from the query image, resulting in the clean query image. }
    \label{fig:arch}
\end{figure*}

\subsection{Matching for Unseen Weather Conditions}
\label{subsec:token}

In \cref{eq:objective}, a weather effect induced by $\omega$ is shared across the given few-shot support set $\mathbf{S}_\omega$ with the query image $\mathbf{X}$, rather than the background $\mathbf{Y}$.
For instance, given two images of rain, the rain streaks are likely to be similar across the images, while the backgrounds may differ. Thus, exploiting the degradation patterns would be a more efficient approach than leveraging the backgrounds in a few-shot weather degraded image restoration.

Inspired by this, we focus on the degradation pattern to 
enable effective utilization of a limited amount of data by leveraging the shared representations.
We formulate the weather-degraded image restoration task as a task that produces a weather degradation pattern $\mathcal{G}(\mathbf{X})$ from an image $\mathbf{X}$:
\begin{equation}
\label{eq:clean_image}
\mathbf{Y}=\mathbf{X} - \mathcal{G}(\mathbf{X}),\quad \mathcal{G}(\mathbf{X}) = \mathbf{P} + (\mathbf{T}^{\circ -1} - 1) \odot (\mathbf{A} - \mathbf{X}),
\end{equation}
where $\mathbf{T}^{\circ -1}$ is the element-wise inverse of $\mathbf{T}$.

To predict $\mathcal{G}(\mathbf{X})$ from $\mathbf{X}$, we employ a matching-based meta-learning framework~\cite{kim2023universal} that predicts a query label via patch-wise matching.
%
%
%
Given a support set $\{(\mathbf{X}_i, \mathbf{Y}_i)\}_{i\le N} = \{(\mathbf{x}^{i}_{k}, \mathbf{y}^{i}_{k})\}_{i\le N, k \le hw}$ and the patch size of $M = h\times w$, 
MetaWeather predicts the degradation pattern of a query image $\mathbf{X}^{Q}\in \mathbb{R}^{H\times W \times 3}$ by,

\begin{equation}
    f(\mathcal{G}(\mathbf{x}^{Q}_j)) = f(\mathbf{x}^{Q}_j - \mathbf{y}^{Q}_j) = \sum_{i=1}^N \sum_{k=1}^{M} \sigma(f(\mathbf{x}^{Q}_j), f(\mathbf{x}^{i}_{k})) f(\mathbf{x}^{i}_{k} - \mathbf{y}^{i}_{k}),
\end{equation}
where $\mathbf{X}^{Q}=\{\mathbf{x}^Q_j\}_{j \le M}$, $f$ is the encoder, and $\sigma$ is a similarity function. We extend the framework by replacing the query label $\mathbf{y}_j^Q$ with the degradation pattern $\mathcal{G}(\mathbf{x}_j^Q)$ and the label of the support set $\mathbf{y}_k^i$ with $\mathbf{x}_k^i-\mathbf{y}_k^i$.
By introducing a decoder $h \approx f^{-1}$, the weather degradation pattern of the query image $\mathcal{G}(\mathbf{X}^Q)$ can be predicted from the encoded embeddings. 


\subsubsection{Spatial-Channel Matching.}
\label{sec:SCM}
We present spatial-channel matching that takes into account the attribute of weather effects. 
\cref{eq:clean_image} decomposes the degradation pattern $\mathcal{G}(\mathbf{X})$ into degradation particles $\mathbf{P}$ and the atmospheric effect $(\mathbf{T}^{\circ -1} - 1) \odot (\mathbf{A} - \mathbf{X})$.
These decomposed traits of weather effects suggest that the matching process should account for both the particles and the atmospheric effects. 
Spatial matching can address particles such as rain streaks, however, it is unsuitable for handling atmospheric effects since they are globally disseminated within an image. As atmospheric effects are often considered as a style and channel-wise features contain more style information compared to spatial features~\cite{ulyanov2016instance,gatys2016image,lee2022fifo}, we utilize channel information to extract atmospheric effects.


Motivated by these aspects, we devise the spatial-channel matching to extract the representation of atmospheric effects via channel-wise matching, while the degradation particles are extracted via spatial matching, in a decomposed manner.
As channel matching would focus primarily on extracting atmospheric effects, spatial matching naturally prioritizes the extraction of degradation particles.
By leveraging both dimensions in the matching module, the model can fully extract the weather representations from a few-shot support set and utilize them to predict the degradation patterns of incoming query images.
%


\subsection{Implementation}
\label{subsec:arch}

\begin{figure}[t]
    \centering
    \includegraphics[width=1.0\linewidth]{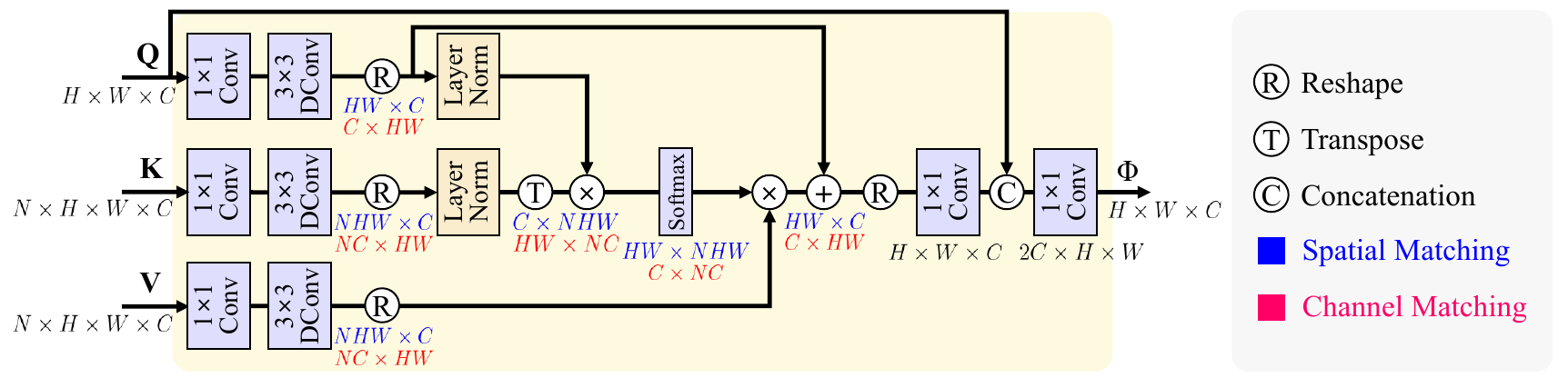}
    \caption{Attention modules in our matching module. 
    }
    \label{fig:mm}
\end{figure}


The architecture of MetaWeather comprises a single-encoder, a single-decoder, and matching modules, as shown in \cref{fig:arch}. The encoder and decoder have hierarchical architectures. The encoder reduces spatial size while going deeper, while the decoder increases spatial size gradually following a U-Net-ike structure~\cite{ronneberger2015u}.

\subsubsection{Encoder.}
The main objective of the encoder is to extract enriched and informative features of the query and the support set images to facilitate more effective matching between query and support images.
These enriched features are passed on to the matching modules, where detailed comparisons and matching between the query and support images take place.
The same encoder is shared to encode query and support set images.
We adopt Swin Transformer~\cite{liu2021swin} as our encoder. 
As the Swin Transformer hierarchically reduces spatial size and enlarges channel capacity, its intact structure can play the role of an encoder in a U-net-like structure, which has shown prominent performances on low-level vision tasks~\cite{wang2022uformer,zamir2022restormer,chen2022simple}.
To build further generalized representation, we utilize SimMIM~\cite{xie2022simmim} pre-trained knowledge to initialize our encoder.




%


\subsubsection{Matching Module.} 

In order to make matching between the query image and support set images, the encoded features are bypassed from encoder blocks to matching modules.
We employ channel and spatial attention in parallel in our matching module for more detailed matching results. The matching module is depicted in \cref{fig:mm}.


At each level of the encoder, given an encoded embedding $\mathbf{Q} \in \mathbb{R}^{H \times W\times C}$ of the query image $\mathbf{X^Q}$, and embeddings $\mathbf{K}, \mathbf{V} \in \mathbb{R}^{N \times H \times W \times C}$ of the support set $\{(\mathbf{X}_i, \mathbf{Y}_i)\}_{i \le N}$, we project the embeddings into matrices $\mathbf{q} \in \mathbb{R}^{d \times l}, \mathbf{k, v} \in \mathbb{R}^{Nd \times l}$ by 
$\mathbf{q}=\mathbf{W}_2^Q \mathbf{W}_1^Q \mathbf{Q}, \mathbf{k}=\mathbf{W}_2^K \mathbf{W}_1^K \mathbf{K}, $ and $\mathbf{v}=\mathbf{W}_2^V \mathbf{W}_1^V \mathbf{V}$, where $\mathbf{W}_1^{(\cdot)}$ and $ \mathbf{W}_2^{(\cdot)}$ are the $1 \times 1$ point-wise and the $3 \times 3$ depth-wise convolution, respectively.
Note that $d = HW$ and $l = C$ for spatial matching and $d = C$ and $l = HW$ for channel matching.
We apply layer normalization to the $\mathbf{q}$ and the $\mathbf{k}$ for effective matching,
$\mathbf{\overline{q}}=\text{LayerNorm}(\mathbf{q}), \mathbf{\overline{k}}=\text{LayerNorm}(\mathbf{k})$. We predict the label of the query $\Phi$ by leveraging a modified Multi-Dconv Head Transposed Attention (MDTA)~\cite{zamir2022restormer},

\begin{align}
    \Phi &= \mathbf{W}_1^\Phi \text{Cat}\big(\text{MDTA}(\mathbf{Q}, \mathbf{K}, \mathbf{V}), \mathbf{Q}\big), \\
    &\text{where MDTA}(\mathbf{Q}, \mathbf{K}, \mathbf{V}) = \mathbf{W}_1^O \big(\text{Softmax}(\mathbf{\overline{q\vphantom{d}}}\,\mathbf{\overline{k}}^\top/\alpha)\mathbf{v} + \mathbf{q} \big),
\end{align}
where $\alpha$ is a learnable temperature factor and Cat($\cdot, \cdot$) is the channel-wise concatenation operator. Consequently, the final output of the matching module is the sum of the outputs of channel and spatial matching modules.

\subsubsection{Decoder.}

The decoder of our model receives results from the matching module for each hierarchical level. 
Subsequently, our decoder constructs the degradation pattern of the query image, utilizing the matched results obtained at each hierarchical level.
After merging all the matched results, the output is further refined by a refinement block and a convolution layer. The refined output is subtracted from $\mathbf{X^Q}$, followed by a final convolution layer.
We adopt the hierarchical decoder architecture presented in Chen \textit{et al.}~\cite{chen2022simple} as our decoder due to its seamless compatibility with our hierarchical architecture. 


\subsubsection{Training Strategy.}
\label{subsec:train}


We employ a conventional episodic meta-learning protocol to obtain and preserve generalized knowledge, following the training strategy of the universal meta-learning framework~\cite{kim2023universal}. We build general knowledge during the meta-train phase and then adapt the model to the given weather condition during the meta-test phase.
Specifically, during the meta-train phase, we imitate the meta-test phase by partitioning batch data into query images and support set images. 
All parameters of our model are trained end-to-end during the meta-train phase.

Once the meta-training is complete, we adapt our architecture to unseen weather types during the meta-test phase. 
In the meta-test phase, we split the support set into two groups: half of the data serves as the query set, while the other half serves as the support set. 
With each iteration, we exchange the roles of the query and support set. 
We train only bias parameters of our model to preserve the knowledge acquired in the meta-train phase.

\section{Experiments}
\label{sec:experiments}


\subsection{Experimental Settings}
\label{subsec:setup}
\par

\par

\subsubsection{Baselines.}
We compare our method with state-of-the-art generic image restoration methods. MPRNet~\cite{zamir2021multi}, Restormer~\cite{zamir2022restormer}, MAXIM~\cite{tu2022maxim}, NAFNet~\cite{chen2022simple}, and FocalNet~\cite{cui2023focal} are initially proposed to be trained independently on individual specific degradation types. However, since they share the same architecture across tasks, they can seamlessly perform on our setting. AirNet~\cite{li2022all} ,TransWeather~\cite{valanarasu2022transweather}, and WeatherDiffusion~\cite{ozdenizci2023weatherdiffusion} are designed to be jointly trained with multiple types of degradation patterns, making their methods suitable to be our baseline models as well. 
Note that \textbf{{all baselines are fine-tuned}} with the same number of support set as ours during the meta-test phase.

\par
\subsubsection{Evaluation Metrics.}
We compare our method with the benchmark methods on PSNR (dB) and SSIM, following 
Zamir \textit{et al.}~\cite{zamir2022restormer} and Valanarasu \textit{et al.}~\cite{valanarasu2022transweather}. Higher values for both metrics indicate improved performance.

\subsubsection{Implementation Details.} 
Our method is implemented on the PyTorch framework, with training and testing performed on an NVIDIA RTX 3090 GPU. We use the Swin Transformer Base (Swin-B) model as the backbone for our encoder. 
We set the numbers of the attention heads in the matching modules as [4,8,16,16], from the top layer to the bottom layer. 
In the meta-train phase, we set the learning rate for the encoder as ${10}^{-5}$ and for the other parts of our model as ${10}^{-4}$. During the meta-test phase, the learning rate for the bias parameters is set to ${10}^{-6}$.
We use the AdamW optimizer~\cite{loshchilov2017decoupled} with $\beta_1 = 0.9$ and $\beta_2 = 0.999$. 
An L1 loss function is used for both the meta-train and meta-test phases. In the meta-train phase, we set the batch size to 8, while the batch size is set to equal the number of few-shot images in the meta-test phase. In all experiments, the number of images in the support set is set to 1 by default.
All images are resized to 224 $\times$ 224 pixels. 
The models are trained for 300k iterations during the meta-training phase and 20k iterations during the meta-test phase.

\begin{table*}[t!]
\centering
\caption{Quantitative comparison on BID Task \RNum{2}.A dataset~\cite{han2022blind}. 
The \textbf{best} and \underline{second-best} results are highlighted.}
\def\arraystretch{1.25}
\resizebox{1.0\columnwidth}{!}{
{
\setlength{\tabcolsep}{1.5pt}
\begin{tabular}{c|cc|cc|cc|cc|cc||cc}
\hline
\toprule
\multirow{2}{*}{Model} & \multicolumn{2}{c|}{$\text{R+S}$} & \multicolumn{2}{c|}{$\text{R+H}_\text{L}$} & \multicolumn{2}{c|}{$\text{R+H}_\text{H}$} & \multicolumn{2}{c|}{R+D+H$_\text{M}$} & \multicolumn{2}{c||}{R+D+S+H$_\text{M}$}  & \multicolumn{2}{c}{Average}\\
\multirow{2}{*}{} & PSNR & SSIM & PSNR & SSIM & PSNR & SSIM & PSNR & SSIM  & PSNR & SSIM & PSNR & SSIM\\
\hline
MPRNet~\cite{zamir2021multi} & 25.96 & 0.8397 & 22.75 & 0.8572 & 18.02 & 0.7291 & 18.70 & 0.7391 & 18.57 & 0.6682 & 20.80 & 0.7667\\
TransWeather~\cite{valanarasu2022transweather} & 24.27 & 0.7869 & 22.89 & 0.8394 & 18.20 & 0.7348 & 19.63 & 0.7365 & 19.45 & 0.6412 & 20.89 & 0.7478\\
AirNet~\cite{li2022all}  & 25.34 & 0.7631 & 21.25 & 0.7704 & 17.00 & 0.6738 & 17.50 & 0.6404 & 18.42 & 0.5908 & 19.90 & 0.6877\\
MAXIM~\cite{tu2022maxim} & 26.44 & 0.8086 & 21.11 & 0.8298 & 17.51 & 0.7054 & 15.49 & 0.6519 & 17.78 & 0.6270 & 19.67 & 0.7245\\
Restormer~\cite{zamir2022restormer} & 26.39 & \underline{0.8501} & 22.81 & 0.8450 & 15.43 & 0.6733 & 19.92 & 0.7602 & 16.75 & 0.6713 & 20.26 & 0.7600\\
NAFNet~\cite{chen2022simple} & \underline{26.82} & 0.8393 & 22.40 & 0.8508 & 17.78 & 0.7207 & 19.87 & 0.7602 & 16.95 & 0.6609 & 20.76 & 0.7664\\
WeatherDiffusion~\cite{ozdenizci2023weatherdiffusion} & 25.54 & 0.8431 & 23.17 & 0.8077 & \underline{21.17} & \underline{0.7720} & \underline{20.64} & 0.7623 & \underline{20.80} & \textbf{0.7735} & \underline{22.26} & \underline{0.7917}\\
FocalNet~\cite{cui2023focal} & 25.85 & 0.8164 & \underline{23.47} & \underline{0.8604} & 19.28 & 0.7566 & 20.46 & \underline{0.7663} & 18.57 & 0.6707 & 21.53 & 0.7741\\
\hline
\textbf{MetaWeather} & \textbf{27.88} & \textbf{0.8574} & \textbf{25.85} & \textbf{0.8790} & \textbf{22.02} & \textbf{0.8137} & \textbf{23.24} & \textbf{0.8142} & \textbf{22.42} & \underline{0.7575} & \textbf{24.28} & \textbf{0.8244}\\
\bottomrule
\end{tabular}
}
}

\label{tab:main_table}
\end{table*}


\subsection{Experiments on Co-occurring Weather Types}
\label{subsec:co-occurring}
\subsubsection{Datasets.}
In the meta-train phase, we train our proposed framework, \linebreak MetaWeather, on a combination of basic weather-type datasets. And in the meta-test phase, we adapt our framework to datasets consisting of composited weather patterns.
As a meta-train dataset, we use multiple datasets each of which includes one of the following degradation patterns: rain, snow, fog, and raindrop. 
The datasets are Rain13K dataset~\cite{zamir2021multi} for rain, Snow100K dataset~\cite{Liu2018desnow} for snow, Foggy Cityscapes~\cite{sakaridis2018semantic} for fog, Raindrop dataset~\cite{qian2018attentive} for raindrop.


We use the BID Task~\RNum{2}.A dataset~\cite{han2022blind} for the meta-test datasets. The dataset has several different combinations of various synthetic weather degradation patterns. As shown in \cref{fig:fig1}, the combination of basic weather types can be considered as a distinct weather type due to the noticeable decrease in the performance of existing methods. 
Each case contains a combination of weather degradation patterns as follows: rain streak + snow (R+S); rain streak + light haze (R+$\text{H}_\text{L}$); rain streak + heavy haze (R+$\text{H}_\text{H}$); rain streak + raindrop + moderate haze (R+D+$\text{H}_\text{M}$); rain streak + raindrop + snow + moderate haze (R+D+S+$\text{H}_\text{M}$).

\subsubsection{Results.}
\cref{tab:main_table} compares the performance of our method with the baselines. 
MetaWeather achieves 2.02 dB performance increase in PSNR and 0.0327 in SSIM on average, compared to the second-best model, WeatherDiffusion.
Although TransWeather, AirNet, and WeatherDiffusion are proposed to deal with multiple weathers by training jointly with multiple datasets, they exhibit limited adaptability toward unseen weather types, whereas ours flexibly adapts to the weather conditions.

\begin{figure*}[t!]
    \centering
    \includegraphics[width=\linewidth]{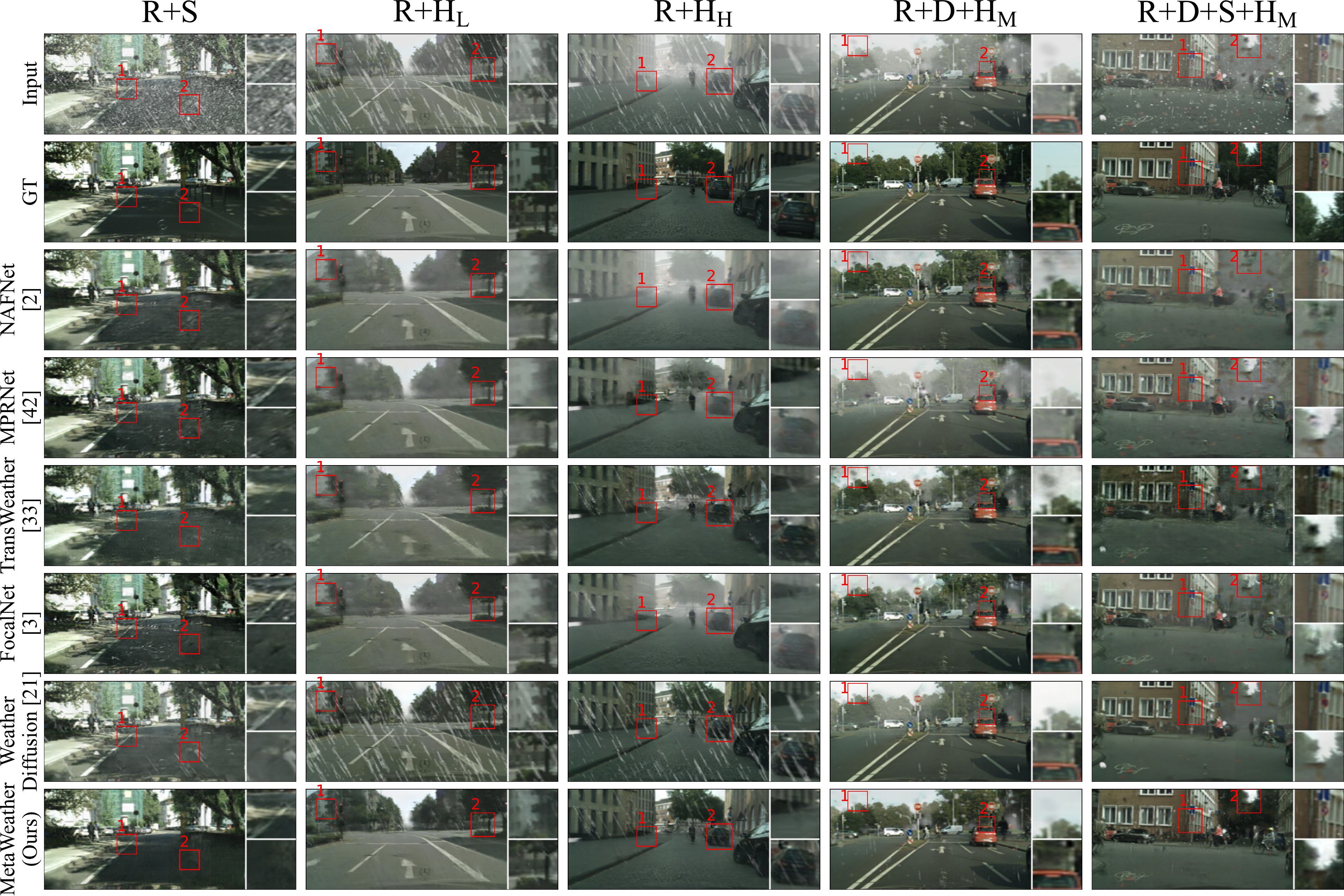}
    \caption{Qualitative comparison on BID Task \RNum{2}.A dataset~\cite{han2022blind}. The results of top five baselines in \cref{tab:main_table} are sampled. Best viewed with zoom and color.}
    \label{fig:main-qual}
\end{figure*}

\cref{fig:main-qual} provides a qualitative comparison with the top five baselines in \cref{tab:main_table}. Despite the mitigation of adverse weather artifacts by other state-of-the-art models, it is evident that degradation patterns are still present. 
In the case of R+S, the baseline models fail to remove the majority of adverse weather-related artifacts, particularly the rain streaks, whereas MetaWeather has removed most artifacts. 
Although WeatherDiffusion achieves the second-best performance in \cref{tab:main_table}, it barely removes the rain streaks and snow particles compared to ours.
Furthermore, as shown in cases of R+H$_\text{H}$, R+D+H$_\text{M}$, and R+D+S+H$_\text{M}$, existing models have limited ability to cope with the hazy effect, while MetaWeather removes almost all degradation patterns.
In conclusion, these results indicate that MetaWeather enables flexible adaptation towards unseen new weather types.

\subsection{Experiments on Real-World Scenario}
\label{subsec:real-world}
\subsubsection{Datasets.}

To further validate the effectiveness of our approach in real-world scenarios, we adapt and evaluate our method to the SPA-Data~\cite{Wang2019spa} (rain+veiling effect) and the RealSnow dataset~\cite{zhu2023learning} (snow+veiling effect), where the veiling effects naturally emerge due to the accumulation of weather patterns. The ground-truth images in these real-world datasets are obtained by applying a median filter over the frames to extract non-degraded parts of the images. 

\begin{table*}[t!]
\centering
\caption{Quantitative comparison on SPA-Data~\cite{Wang2019spa} and RealSnow~\cite{zhu2023learning} dataset. $\dagger$ denotes a method for real-world weather-degraded image restoration. The \textbf{best} results are highlighted.}
\def\arraystretch{1.15}
\resizebox{0.7\columnwidth}{!}
{
{
\setlength{\tabcolsep}{5pt}
\begin{tabular}{c|cc|cc}
\hline
\toprule
\normalfont{ } 
\multirow{2}{*}{Model} & \multicolumn{2}{c|}{SPA-Data~\cite{wang2019panet}}  & \multicolumn{2}{c}{RealSnow\cite{zhu2023learning}}\\
 & PSNR & SSIM & PSNR & SSIM\\
\hline
MPRNet~\cite{zamir2021multi} & 26.57 & 0.8355 & 26.84 & 0.8773\\
TransWeather~\cite{valanarasu2022transweather} & 26.62 & 0.8508 & 24.38 & 0.8354\\
AirNet~\cite{li2022all} & 27.72 & 0.8430 & 21.69 & 0.8244\\
MAXIM~\cite{tu2022maxim} & 32.21 & 0.9002 & 27.30 & 0.8930\\
Restormer~\cite{zamir2022restormer} & 31.18 & 0.9089 & 29.65 & 0.9037\\
NAFNet~\cite{chen2022simple} & 32.16 & 0.9144 & 28.19 & 0.9014\\
WeatherDiffusion~\cite{ozdenizci2023weatherdiffusion} & 31.18 & 0.9014 & 26.32 & 0.8779\\
FocalNet~\cite{cui2023focal} & 29.03 & 0.8846 & 24.81 & 0.8527\\
Patil \textit{et al.}$^{\dagger}$~\cite{patil2023multi} & 20.26 & 0.8291 & 20.22 & 0.8974\\
\hline
\textbf{MetaWeather (Ours)} & \textbf{34.81} & \textbf{0.9394} & \textbf{30.15} & \textbf{0.9222}\\
\bottomrule
\end{tabular}
}
}

\label{tab:real_world}
\end{table*}

\subsubsection{Baselines.} In the real-world evaluation, we include the method proposed by Patil \textit{et al.}~\cite{patil2023multi} alongside the baselines used in \cref{subsec:co-occurring}. This method is intended for utilization in real-world weather-degraded image restoration only with pre-trained knowledge. 
Consequently, we assess this method in our real-world experiments without additional adaptation.

\begin{figure*}[t!]
    \centering
    \includegraphics[width=\linewidth]{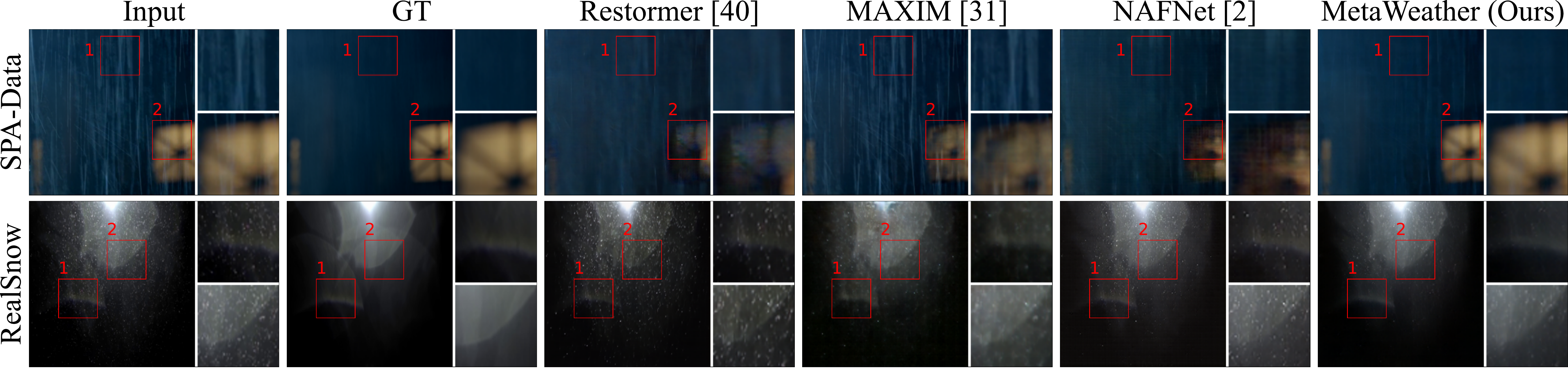}
    \caption{Qualitative comparison on SPA-Data~\cite{Wang2019spa} and RealSnow~\cite{zhu2023learning} dataset. The results of top three baselines in \cref{tab:real_world} are sampled. Best viewed with zoom and color.}
    \label{fig:real-qual}
\end{figure*}

\subsubsection{Results.}
As shown in \cref{tab:real_world}, our method outperforms the baselines, similar to the previous results on unseen weather types, demonstrating its effectiveness in real-world scenarios as well. Surprisingly, although Patil \textit{et al.}~\cite{patil2023multi} is proposed for real-world weather degraded image restoration task, their method exhibits much lower performance on both datasets compared to the baselines and our approach, indicating the necessity of adaptation.

As depicted in \cref{fig:real-qual}, We qualitatively compare ours with the top three baselines in \cref{tab:real_world}. While the baselines barely remove the rain streaks and snow particles, ours removes them almost perfectly. Additionally, while others distort the background parts, ours preserves the background, only removing the degradation patterns.

\subsection{Ablation Study}
\begin{table}[t!]
\centering
\caption{Ablation study on matched features in matching modules. The \textbf{best} results are highlighted.}
\def\arraystretch{1.3}
{
\resizebox{1\columnwidth}{!}{
\begin{tabular}{c|cc|cc|cc|cc|cc||cc}
\hline
\toprule
\multirow{2}{*}{Matched features}  & \multicolumn{2}{c|}{$\text{R+S}$} & \multicolumn{2}{c|}{$\text{R+H}_\text{L}$} & \multicolumn{2}{c|}{$\text{R+H}_\text{H}$} & \multicolumn{2}{c|}{R+D+H$_\text{M}$} & \multicolumn{2}{c||}{R+D+S+H$_\text{M}$}  & \multicolumn{2}{c}{Average}\\
\multirow{2}{*}{} & PSNR & SSIM & PSNR & SSIM & PSNR & SSIM & PSNR & SSIM  & PSNR & SSIM & PSNR & SSIM\\
\hline
Background & 26.89 & 0.8220 & 23.41 & 0.8181 & 20.11 & 0.7436 & 21.23 & 0.7440 & 18.92 & 0.6567 & 22.11 & 0.7569\\
Degradation pattern &  \textbf{27.88} & \textbf{0.8574} & \textbf{25.85} & \textbf{0.8790} & \textbf{22.02} & \textbf{0.8137} & \textbf{23.24} & \textbf{0.8142} & \textbf{22.42} & \textbf{0.7575} & \textbf{24.28} & \textbf{0.8244}\\

\bottomrule
\end{tabular}
}
}

\label{tab:contents}
\end{table}

\label{subsec:ablation}

\subsubsection{Matched features in matching modules.}
\cref{tab:contents} shows the effect of the matched features within matching modules. We maintain the same architecture and training strategy for this experiment but change the matched features in the matching module to the background of the query image.

Matching the degradation patterns significantly outperforms matching the background, with 2.17 dB in PSNR and 0.0675 in SSIM on average. This is because the predominant common factor across the few-shot sets is the degradation pattern, not the background, as mentioned in~\cref{subsec:token}. In conclusion, learning to match the degradation pattern makes it much easier to generalize the representation of new weather effects compared to matching the background in few-shot weather degraded image restoration tasks.

\subsubsection{Spatial-channel matching.}
\label{sec:exp:SCM}
\begin{table*}[t!]
\centering
\caption{Ablation study on spatial-channel matching. The \textbf{best} and \underline{second-best} results are highlighted. }
\def\arraystretch{1.25}
\resizebox{1.0\columnwidth}{!}
{
\setlength{\tabcolsep}{1.5pt}
\begin{tabular}{cc|cc|cc|cc|cc|cc||cc}
\hline
\toprule
\normalfont{ } 
\multirow{2}{*}{\makecell{Spatial\\matching}} & \multirow{2}{*}{\makecell{Channel\\\;matching\;}}  & \multicolumn{2}{c|}{$\text{R+S}$} & \multicolumn{2}{c|}{$\text{R+H}_\text{L}$} & \multicolumn{2}{c|}{$\text{R+H}_\text{H}$} & \multicolumn{2}{c|}{R+D+H$_\text{M}$} & \multicolumn{2}{c||}{R+D+S+H$_\text{M}$}  & \multicolumn{2}{c}{Average}\\
\multirow{2}{*}{} & \multirow{2}{*} & PSNR & SSIM & PSNR & SSIM & PSNR & SSIM & PSNR & SSIM  & PSNR & SSIM & PSNR & SSIM\\
\hline
- & - & 27.12 & 0.8369 & \underline{24.79} & 0.8534 & 20.06 & 0.7611 & 21.41 & 0.7613 & 20.71 & 0.6997 & 22.82 & 0.7825\\
\checkmark & - &  27.23 & 0.8282 & 23.75 & 0.8314 & 20.41 & 0.7675 & 22.33 & 0.7798 & 21.52 & 0.7088 & 23.05 & 0.7831\\
 - & \checkmark &  \underline{27.85} & \underline{0.8570} & {24.66} & \underline{0.8696} & \textbf{22.17} & \underline{0.8098} & \underline{22.52} & \underline{0.8012} & \underline{22.31} & \underline{0.7529} & \underline{23.90} & \underline{0.8181}\\
 \checkmark & \checkmark & \textbf{27.88} & \textbf{0.8574} & \textbf{25.85} & \textbf{0.8790} & \underline{22.02} & \textbf{0.8137} & \textbf{23.24} & \textbf{0.8142} & \textbf{22.42} & \textbf{0.7575} & \textbf{24.28} & \textbf{0.8244}\\

\bottomrule
\end{tabular}
}

\label{tab:ablation_table}
\end{table*}

\cref{tab:ablation_table} shows the impact of spatial and channel-wise matching in the matching modules. The use of matching mechanisms facilitates flexible adaptation in few-shot weather degraded image restoration tasks. The performance gap between the models increases as degradation patterns accumulate, suggesting that matching plays a key role in flexible adaptation. 

However, spatial matching shows a limited performance increase, which is similar to the framework~\cite{kim2023universal} we extend, showing a lack in restoring the veiling effects. Using channel matching outperforms using spatial matching solely, implying that leveraging channel-wise information is crucial in weather degraded image restoration tasks. Moreover, when both spatial and channel matching are used together, the performance improves further, indicating that matching in both dimensions is beneficial for learning the weather effect.

\begin{figure*}[t!]
    \centering
    \includegraphics[width=\linewidth]{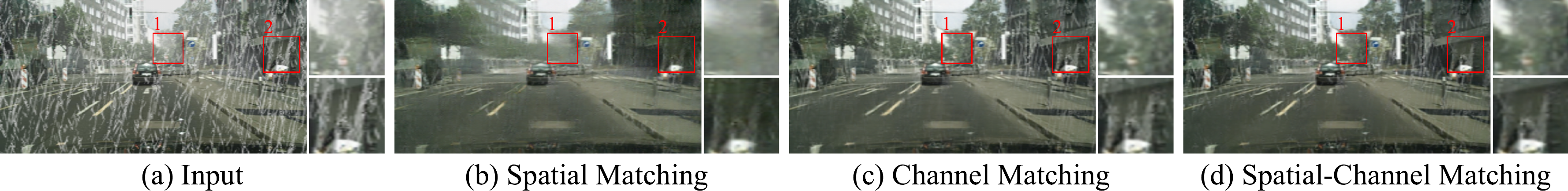}
    \caption{Qualitative comparison on the effect of spatial-channel matching. Best viewed with zoom and color.} 
    \label{fig:spatial-channel matching}
\end{figure*}

\cref{fig:spatial-channel matching} illustrates the effectiveness of spatial and channel-wise matching. For qualitative comparison on spatial-channel matching, we evaluate the variants of matching modules on the R+H$_\text{L}$ weather type in BID Task \RNum{2}.A dataset.
\cref{fig:spatial-channel matching} (b) shows that using only spatial matching restores the rain streaks, while foggy degradation patterns still remain. Conversely, \cref{fig:spatial-channel matching} (c) illustrates that channel matching effectively removes the fog effect while the rain streaks still remain. 
Lastly, \cref{fig:spatial-channel matching} (d) depicts that utilizing spatial and channel matching together removes both degradation patterns. This result implies that matching in each dimension effectively handles the weather effects in a decomposed manner, as described in \cref{sec:SCM}.

\subsubsection{The number of shots in the support set.}

\begin{table*}[t]
\centering
\caption{Ablation study on the number of shots in the support set. The \textbf{best} and \underline{second-best} results for each number of shots are highlighted.}
\def\arraystretch{1.25}
\resizebox{1.0\columnwidth}{!}{
{
\begin{tabular}{c|c|cc|cc|cc|cc|cc||cc}
\hline
\toprule
\normalfont{ } 
\multirow{2}{*}{\makecell{\# of\\\;shots\;}} & \multirow{2}{*}{Model}  & \multicolumn{2}{c|}{$\text{R+S}$} & \multicolumn{2}{c|}{$\text{R+H}_\text{L}$} & \multicolumn{2}{c|}{$\text{R+H}_\text{H}$} & \multicolumn{2}{c|}{R+D+H$_\text{M}$} & \multicolumn{2}{c||}{R+D+S+H$_\text{M}$}  & \multicolumn{2}{c}{Average}\\
\multirow{2}{*}{} & \multirow{2}{*}{} & PSNR & SSIM & PSNR & SSIM & PSNR & SSIM & PSNR & SSIM & PSNR & SSIM & PSNR & SSIM \\
\hline
\multirow{3}{*}{1}& WeatherDiffusion~\cite{ozdenizci2023weatherdiffusion} & 25.54 & \underline{0.8432} & 23.12 & 0.8066 & \underline{21.17} & \underline{0.7724} & \underline{20.70} & 0.7646 & \underline{20.81} & \textbf{0.7734} & \underline{22.27} & \underline{0.7920}\\
\multirow{3}{*}{}& FocalNet~\cite{cui2023focal} & \underline{25.85} & 0.8164 & \underline{23.47} & \underline{0.8604} & 19.28 & 0.7566 & 20.46 & \underline{0.7663} & 18.57 & 0.6707 & 21.53 & 0.7741\\
\multirow{3}{*}{}& \textbf{MetaWeather (Ours)} & \textbf{27.88} & \textbf{0.8574} & \textbf{25.85} & \textbf{0.8790} & \textbf{22.02} & \textbf{0.8137} & \textbf{23.24} & \textbf{0.8142} & \textbf{22.42} & \underline{0.7575} & \textbf{24.28} & \textbf{0.8244}\\
\hline
\multirow{3}{*}{2}& WeatherDiffusion~\cite{ozdenizci2023weatherdiffusion}  & 25.54 & \underline{0.8431} & 23.17 & 0.8077 & \underline{21.17} & \underline{0.7720} & \underline{20.64} & \underline{0.7623} & \underline{20.80} & \underline{0.7735} & \underline{22.26} & \underline{0.7917}\\
\multirow{3}{*}{}& FocalNet~\cite{cui2023focal}  & \underline{26.58} & 0.8358 & \underline{24.51} & \underline{0.8636} & 20.16 & 0.7677 & 20.62 & 0.7599 & 18.79 & 0.6909 & 22.13 & 0.7836\\
\multirow{3}{*}{}& \textbf{MetaWeather (Ours)} & \textbf{28.54} & \textbf{0.8728} & \textbf{25.77} & \textbf{0.8876} & \textbf{22.58} & \textbf{0.8281} & \textbf{22.80} & \textbf{0.8188} & \textbf{22.16} & \textbf{0.7775} & \textbf{24.37} & \textbf{0.8370}\\
\hline
\multirow{3}{*}{4}& WeatherDiffusion~\cite{ozdenizci2023weatherdiffusion} 	& 25.49 & \underline{0.8428} & 23.18 & 0.8077 & \underline{21.19} & 0.7720 & 20.60 & 0.7616 & \underline{20.80} & \underline{0.7743} & 22.25 & \underline{0.7917}\\
& FocalNet~\cite{cui2023focal} 	& \underline{26.46} & 0.8154 & \underline{24.35} & \underline{0.8719} & 19.76 & \underline{0.7757} & \underline{20.93} & \underline{0.7815} & 20.76 & 0.7097 & \underline{22.45} & 0.7908\\
& \textbf{MetaWeather (Ours)} & \textbf{28.62} & \textbf{0.8780} & \textbf{26.17} & \textbf{0.8928} & \textbf{23.40} & \textbf{0.8422} & \textbf{23.59} & \textbf{0.8349} & \textbf{22.99} & \textbf{0.7937} & \textbf{24.95} & \textbf{0.8483}\\

\bottomrule
\end{tabular}
}
}

\label{tab:shot}
\end{table*}

We vary the number of shots in the support set used in the adaptation phase, as presented in \cref{tab:shot}. 
We compare our method with the top three models in \cref{tab:main_table}. 
Our method achieves the best average PSNR and SSIM regardless of the number of support set, indicating that our method is more effective at utilizing and generalizing knowledge extracted from a small number of images. 
Also, as the number of support set increases, the performance gap between our method and the others increases, while other baselines show limited performance improvements.
This demonstrates the scalability and flexibility of our approach compared to baselines.


\subsubsection{Robustness on support set selection.}
\begin{table}[t!]
\centering
\caption{Robustness on the selection of the support set. 
All performance numbers are averaged across cases in the BID Task \RNum{2}.A dataset for each selected support set.}
\def\arraystretch{1.1}
\resizebox{0.8\columnwidth}{!}
{\small
\setlength{\tabcolsep}{4pt}
\begin{tabular}{c|ccccc|cc}
\hline
\toprule
Support set & \# 1 & \# 2 & \# 3 & \# 4 & \# 5 & Mean & Std.   \\
\hline & \\[-2.4ex]
PSNR & 24.28 & 24.35 & 23.83 & 24.20 & 24.12 & 24.16 & 0.20\\ 
SSIM & 0.8244 & 0.8265 & 0.8226 & 0.8256 & 0.8242 & 0.8247 & 0.0018\\
\bottomrule
\end{tabular}
}

\label{tab:support_set}
\end{table}
Since our method utilizes a given support set during adaptation and inference, we verify whether our method is robust to the selection of the support set. 
To probe the robustness, we conduct the same experiments using five randomly selected support sets on all unseen weather types in BID Task \RNum{2}.A dataset.
\cref{tab:support_set} clearly shows that our method is robust to the selection of the support set, with tiny standard deviations in both PSNR and SSIM. 
This robustness stems from our method's ability to effectively extract and generalize representations obtained from a few-shot support set.

\subsubsection{Bias parameter tuning.}

\begin{table*}[t]
\centering
\caption{Fairness on bias parameter tuning. The \textbf{best} results are highlighted.}
\def\arraystretch{1.25}
\resizebox{1.0\columnwidth}{!}{
{
\begin{tabular}{c|cc|cc|cc|cc|cc||cc}
\hline
\toprule
\normalfont{ } 
\multirow{2}{*}{Model}  & \multicolumn{2}{c|}{$\text{R+S}$} & \multicolumn{2}{c|}{$\text{R+H}_\text{L}$} & \multicolumn{2}{c|}{$\text{R+H}_\text{H}$} & \multicolumn{2}{c|}{R+D+H$_\text{M}$} & \multicolumn{2}{c||}{R+D+S+H$_\text{M}$}  & \multicolumn{2}{c}{Average}\\
\multirow{2}{*}{} & PSNR & SSIM & PSNR & SSIM & PSNR & SSIM & PSNR & SSIM & PSNR & SSIM & PSNR & SSIM \\
\hline
WeatherDiffusion~\cite{ozdenizci2023weatherdiffusion} & 25.56 & 0.8434 & 23.18 & 0.8079 & 21.21 & 0.7728 & 20.62 & 0.7619 & 20.81 & 0.7743 & 22.28 & 0.7921\\

FocalNet~\cite{cui2023focal} & 26.18 & 0.8079 & 22.38 & 0.8413 & 17.47 & 0.7271 & 18.39 & 0.7343 & 18.69 & 0.6397 & 20.62 & 0.7501\\

\textbf{MetaWeather} & \textbf{27.88} & \textbf{0.8574} & \textbf{25.85} & \textbf{0.8790} & \textbf{22.02} & \textbf{0.8137} & \textbf{23.24} & \textbf{0.8142} & \textbf{22.42} & \textbf{0.7575} & \textbf{24.28} & \textbf{0.8244}\\

\bottomrule
\end{tabular}
}
}

\label{tab:param_baseline}
\end{table*}

For a fairer comparison, we fine-tune the bias parameters of the baselines in the meta-test phase, following the same adaptation process as ours. We evaluate the top three models in \cref{tab:main_table}. As shown in \cref{tab:param_baseline}, our method remarkably outperforms the baselines with large performance gaps. Specifically, WeatherDiffusion shows a limited increase in performance compared to tuning all parameters, and FocalNet even shows a decrease in performance compared to its results in \cref{tab:main_table}. This is because these models lack flexibility and generalized representation, necessary for flexible adaptation.


\section{Conclusion}
\label{sec:conclusions}

This paper presents MetaWeather, a few-shot learning method capable of restoring images degraded by unseen weather conditions.
To fully exploit the given few-shot support set, we extend the matching-based meta-learning framework to predict weather degradation patterns with a novel matching algorithm.
We guide our model to match the degradation patterns instead of the backgrounds, considering that the degradation patterns are more common across a weather than the backgrounds.
We devise spatial-channel matching to effectively extract the representation from the support images, reflecting the decomposable attribute of the weather effects.
Extensive experiments on co-occurring and real-world unseen weather types show that our method effectively handles unseen weather conditions with a few labeled samples, demonstrating its flexibility and applicability towards the real world. 

%
%
%

%


\putbib
\end{bibunit}

%
%
\clearpage
\begin{bibunit}  
\appendix
\authorrunning{Y. Kim et al.}%
\maketitlesupplementaryfull




\setcounter{section}{0}
\setcounter{table}{0}
\setcounter{figure}{0}
\setcounter{equation}{0}
\renewcommand{\thesection}{\Alph{section}}
\renewcommand{\thetable}{\Alph{table}}
\renewcommand{\thefigure}{\Alph{figure}}
\renewcommand{\theenumi}{\;\;\Alph{enumi}}
\renewcommand{\theequation}{\Alph{equation}}

    
    

\section{Further Experimental Results}
\subsection{Further Ablation Studies}
\subsubsection{Comparison with Visual Token Matching. }

To further verify our proposed method compared to the universal meta-learning framework for dense prediction that we extend, named Visual Token Matching (VTM)~\cite{kim2023universal}, we conduct experiments on co-occurring weather types and real-world scenarios, as in the main paper.
For a fair comparison, we use the same encoder and decoder as ours to tailor the VTM to the task of weather-degraded image restoration. The matching modules only perform spatial matching of the background features to produce the background of the query image.

\cref{tab:vtm} demonstrates the effectiveness of our proposed method compared to the naive application of the VTM for the restoration task. Our method significantly surpasses VTM in all cases due to the matching of degradation patterns instead of the background and the channel-wise matching. The consistent improvement of the performance suggest that the proposed methods help the model to fully exploit the representation of new weather effects.

\begin{figure*}[t!]
    \centering
    \includegraphics[width=\linewidth]{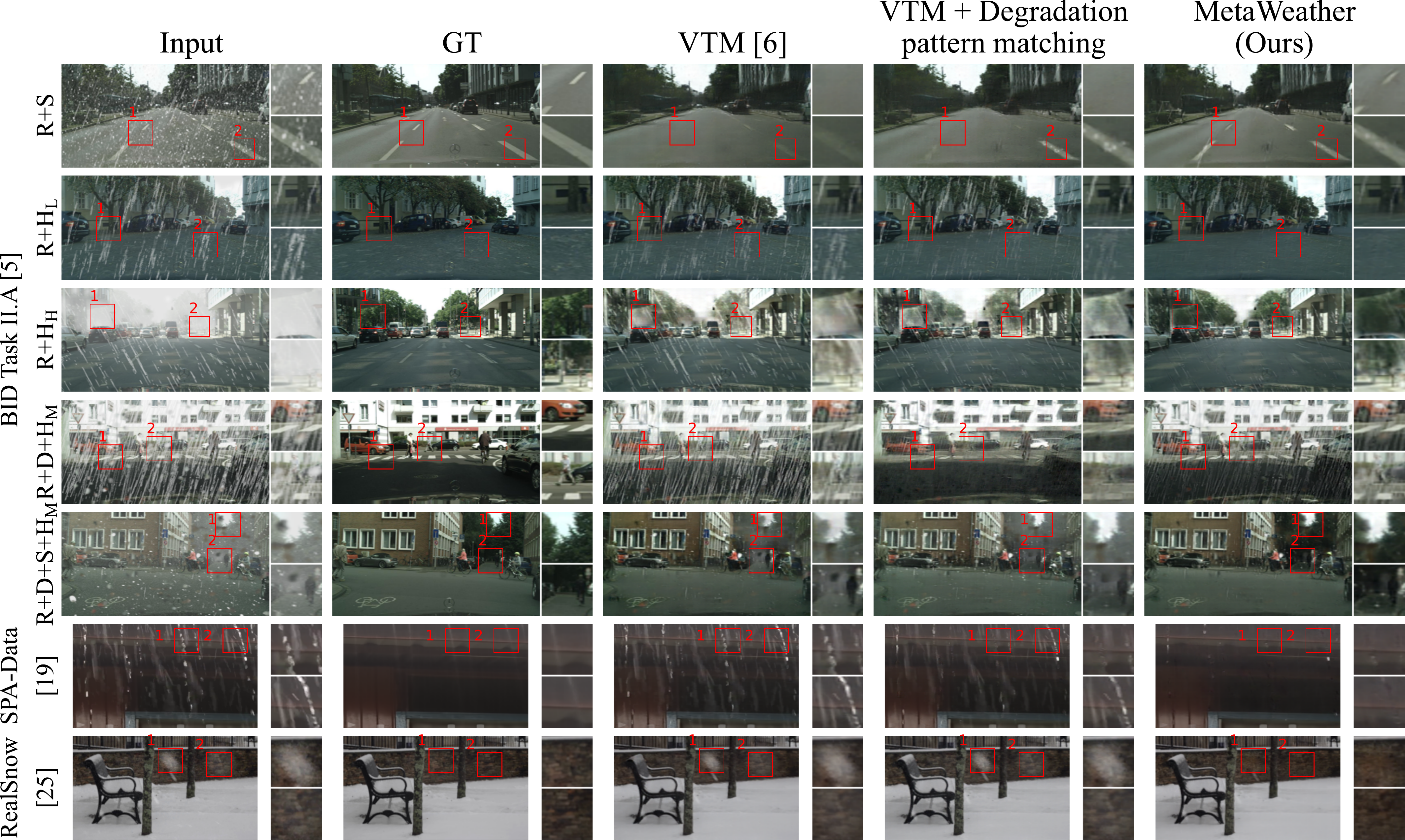}
    \caption{Qualitative comparison with respect to the VTM~\cite{kim2023universal} on the BID Task \RNum{2}.A~\cite{han2022blind}, SPA-Data~\cite{Wang2019spa}, and RealSnow~\cite{Zhang2021desnow} datasets. 
    Best viewed with zoom and color.}
    \label{fig:suppl-vtm-qual}
\end{figure*}

In addition, we provide a qualitative comparison in \cref{fig:suppl-vtm-qual}. The results show that our method effectively restores images while preserving background objects. In contrast, VTM tends to eliminate and distort the background objects, such as lanes, while removing particles. VTM with the degradation pattern matching removes more particles while preserving some background objects, but it still eliminates some objects and leaves fog effects.

\begin{table}[t]
\caption{Ablation study with respect to the VTM~\cite{kim2023universal}. The \textbf{best} results are highlighted.}

\begin{subtable}[h]{\textwidth}
\centering
\def\arraystretch{1.3}
{
\setlength{\tabcolsep}{2pt}
\caption{\small Co-occuring weather types on the BID Task \RNum{2}.A dataset~\cite{han2022blind}.}
\resizebox{1\columnwidth}{!}{
\begin{tabular}{@{}l|cc|cc|cc|cc|cc||cc}
\hline
\toprule
\normalfont{ } 
\multirow{2}{*}{Method}  & \multicolumn{2}{c|}{$\text{R+S}$} & \multicolumn{2}{c|}{$\text{R+H}_\text{L}$} & \multicolumn{2}{c|}{$\text{R+H}_\text{H}$} & \multicolumn{2}{c|}{R+D+H$_\text{M}$} & \multicolumn{2}{c||}{R+D+S+H$_\text{M}$}  & \multicolumn{2}{c}{Average}\\
&  PSNR & SSIM & PSNR & SSIM & PSNR & SSIM & PSNR & SSIM  & PSNR & SSIM & PSNR & SSIM\\
\hline
\ \ VTM~\cite{kim2023universal}  & 26.99 & 0.8275 & 24.15 & 0.8310 & 19.15 & 0.7309 & 20.65 & 0.7388 & 20.03 & 0.6821 & 22.19 & 0.7621\\
\quad  + Degradation pattern matching &  27.23 & 0.8282 & 23.75 & 0.8314 & 20.41 & 0.7675 & 22.33 & 0.7798 & 21.52 & 0.7088 & 23.05 & 0.7831\\
\quad  + Channel matching (\textbf{MetaWeather}) & \textbf{27.88} & \textbf{0.8574} & \textbf{25.85} & \textbf{0.8790} & \textbf{22.02} & \textbf{0.8137} & \textbf{23.24} & \textbf{0.8142} & \textbf{22.42} & \textbf{0.7575} & \textbf{24.28} & \textbf{0.8244}\\

\bottomrule
\end{tabular}
}
}

\label{tab:vtm_bid}
\end{subtable}

\vspace{1em}

\begin{subtable}[hb]{\textwidth}
\centering
\caption{\small Real-world conditions on the SPA-Data~\cite{Wang2019spa} and RealSnow~\cite{zhu2023learning} datasets.}

\def\arraystretch{1.2}
\resizebox{0.7\columnwidth}{!}
{
{
\setlength{\tabcolsep}{5pt}
\begin{tabular}{@{}l|cc|cc}
\hline
\toprule
\normalfont{ } 
\multirow{2}{*}{Method} & \multicolumn{2}{c|}{SPA-Data~\cite{wang2019panet}}  & \multicolumn{2}{c}{RealSnow\cite{zhu2023learning}}\\
 & PSNR & SSIM & PSNR & SSIM\\
\hline
\ \ VTM~\cite{kim2023universal} & 32.27 & 0.9151 & 28.87 & 0.9099\\
\quad + Degradation pattern matching & 33.40 & 0.9234 & 29.85 & \textbf{0.9238} \\
\quad + Channel matching (\textbf{MetaWeather}) & \textbf{34.81} & \textbf{0.9394} & \textbf{30.15} & {0.9222}\\
\bottomrule
\end{tabular}
}
}
\label{tab:vtm_real}
\end{subtable}

\label{tab:vtm}
\end{table}

\subsubsection{Parameters trained in the meta-test phase. }

To further verify the effectiveness of the parameters trained in the meta-test phase, we diversify the parameters trained in the meta-test phase by adopting a different parameter-efficient adaptation method.
Specifically, we change the parameters to layer normalization layers~\cite{qi2022parameter} and all parameters.

\cref{tab:param} shows that updating the bias parameters in the meta-test phase, following the training strategy of the matching-based meta-learning framework~\cite{kim2023universal} we extend, is the most effective in all cases.

\begin{table}[t]
\centering
\caption{Ablation study on parameters trained in the meta-test phase. The \textbf{best} results are highlighted.}
\def\arraystretch{1.5}
\resizebox{1.0\columnwidth}{!}{
{\small
\setlength{\tabcolsep}{2pt}
\begin{tabular}{c|cc|cc|cc|cc|cc||cc}
\hline
\toprule
\normalfont{ } 
\multirow{2}{*}{Trained parameter} & \multicolumn{2}{c|}{$\text{R+S}$} & \multicolumn{2}{c|}{$\text{R+H}_\text{L}$} & \multicolumn{2}{c|}{$\text{R+H}_\text{H}$} & \multicolumn{2}{c|}{R+D+H$_\text{M}$} & \multicolumn{2}{c||}{R+D+S+H$_\text{M}$}  & \multicolumn{2}{c}{Average}\\
\multirow{2}{*}{} & PSNR & SSIM & PSNR & SSIM & PSNR & SSIM & PSNR & SSIM  & PSNR & SSIM & PSNR & SSIM\\
\hline
All & 25.06 & 0.8159 & 23.95 & 0.8504 & 20.62 & 0.7824 & 21.01 & 0.7801 & 20.30 & 0.7161 & 22.19 & 0.7890\\
Layer Normalization~\cite{qi2022parameter}  & 27.70 & 0.8455 & 24.33 & 0.8590 & 20.92 & 0.7851 & 22.28 & 0.7859 & 21.81 & 0.7377 & 23.41 & 0.8026\\

Bias~\cite{zaken2021bitfit} & \textbf{27.88} & \textbf{0.8574} & \textbf{25.85} & \textbf{0.8790} & \textbf{22.02} & \textbf{0.8137} & \textbf{23.24} & \textbf{0.8142} & \textbf{22.42} & \textbf{0.7575} & \textbf{24.28} & \textbf{0.8244}\\
\bottomrule
\end{tabular}
}
}
\label{tab:param}
\end{table}


\clearpage
\subsection{Data efficiency}

We compare the data efficiency of the proposed method with the FocalNet~\cite{cui2023focal}, the third-best baseline from Tab. 1 in the main paper. We exclude the second-best baseline due to its lack of scalability, as shown in Tab. 5 of the main paper.

\begin{figure*}[t]
    \centering
    \includegraphics[width=\linewidth]{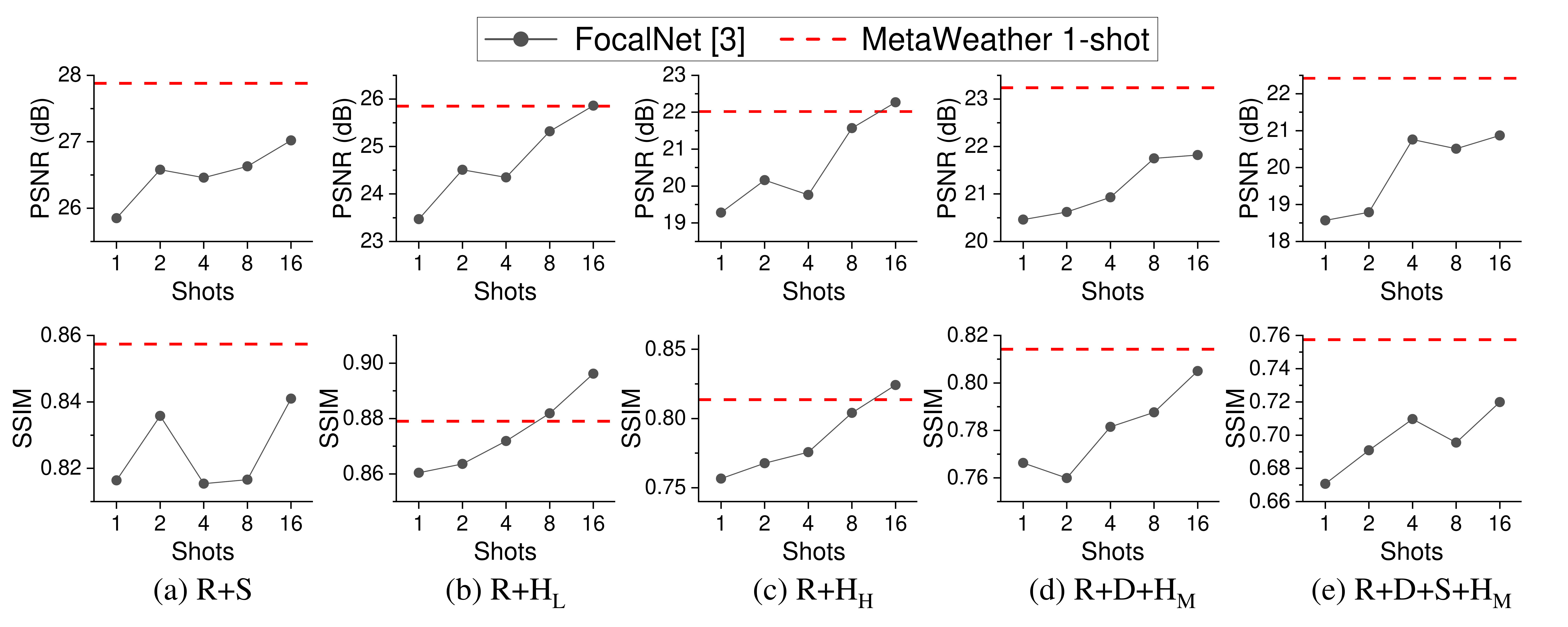}
    \caption{Data efficiency compared with FocalNet~\cite{cui2023focal}.}
    \label{fig:suppl-shots}
\end{figure*}

\cref{fig:suppl-shots} illustrates the data efficiency of MetaWeather. MetaWeather is a practical solution due to its adaptability and data efficiency, especially when faced with the challenge of obtaining labeled data. In most cases, even with a single support pair, our method outperforms the baseline adapted with 16 support pairs.
Notably, the performance gap becomes more pronounced for more complex weather scenarios such as (d) R+D+H$_\text{M}$ and (e) R+D+S+H$_\text{M}$.

\subsection{Additional Qualitative Comparisons}
\begin{figure*}[t]
    \centering
    \includegraphics[width=\linewidth]{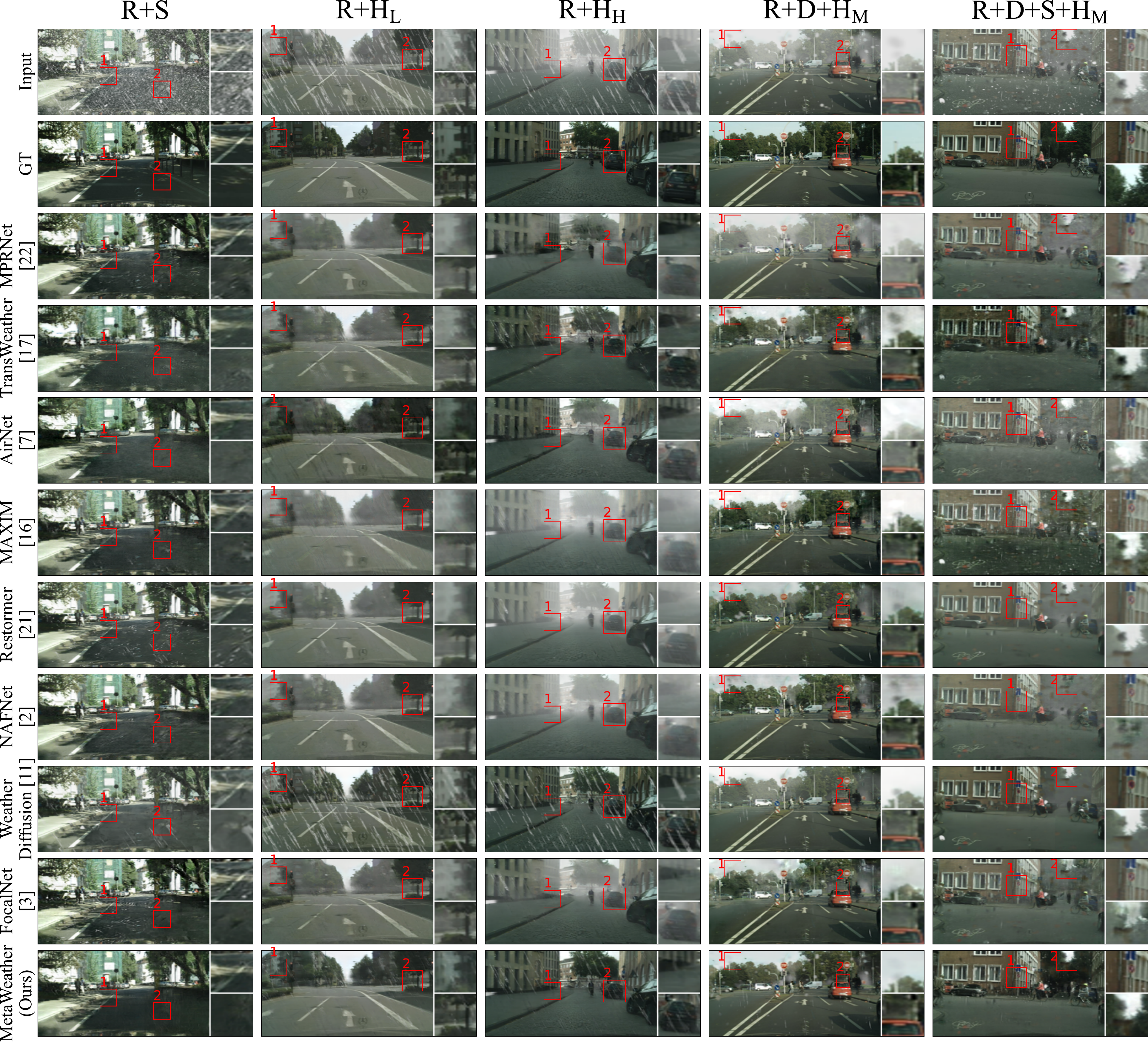}
    \caption{Qualitative comparison on the BID Task \RNum{2}.A dataset~\cite{han2022blind}. 
    Best viewed with zoom and color.}
    \label{fig:suppl-main-qual}
\end{figure*}

\begin{figure*}[ht!]
    \centering
    \begin{subfigure}{\linewidth}
        \centering
        \includegraphics[width=\linewidth]{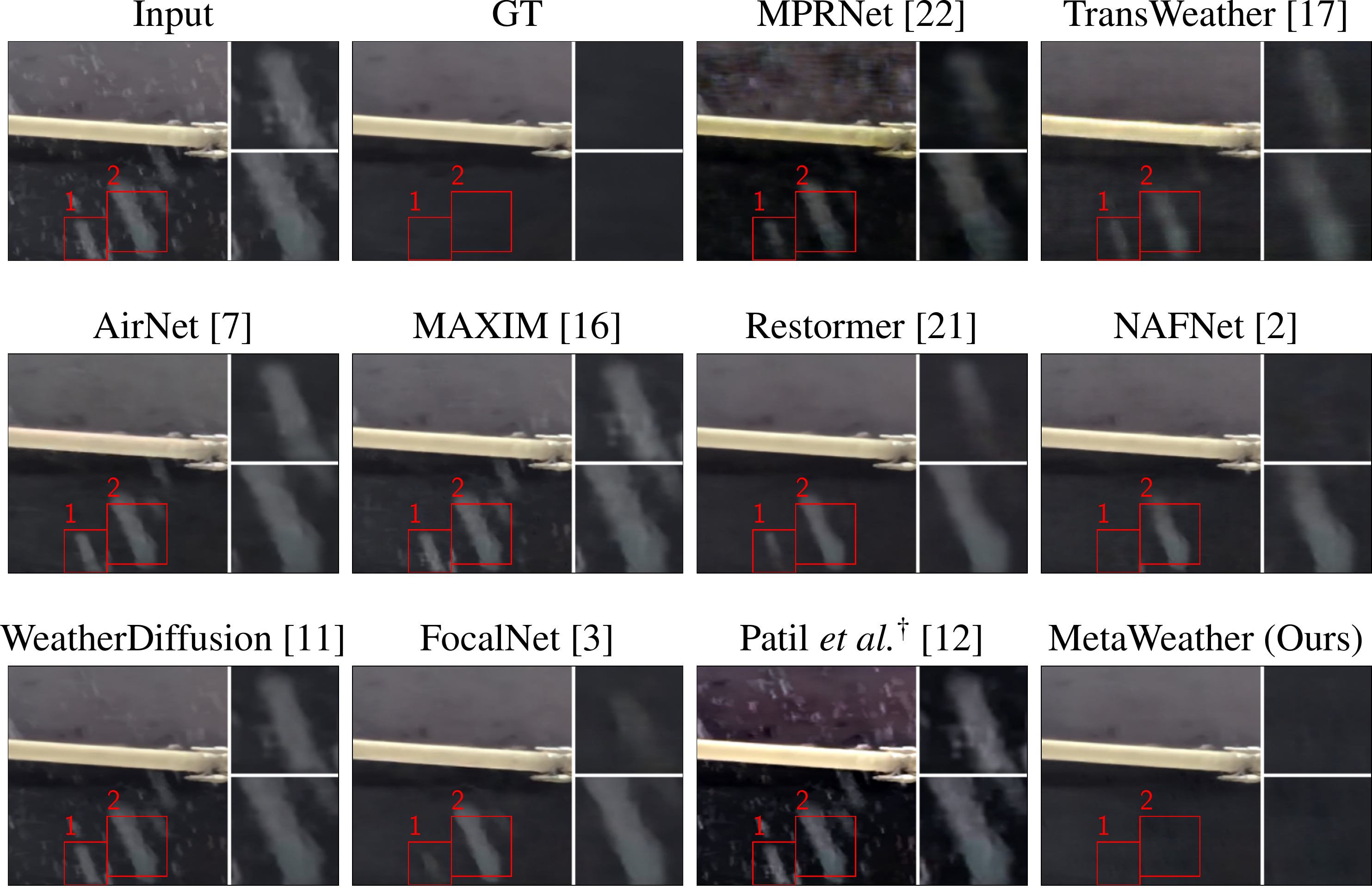}
        \label{fig:suppl-qual-real-spa}
        \caption{SPA-Data~\cite{Wang2019spa}
        \vspace{1em}
        }
    \end{subfigure}
    \begin{subfigure}{\linewidth}
        \centering
        \includegraphics[width=\linewidth]{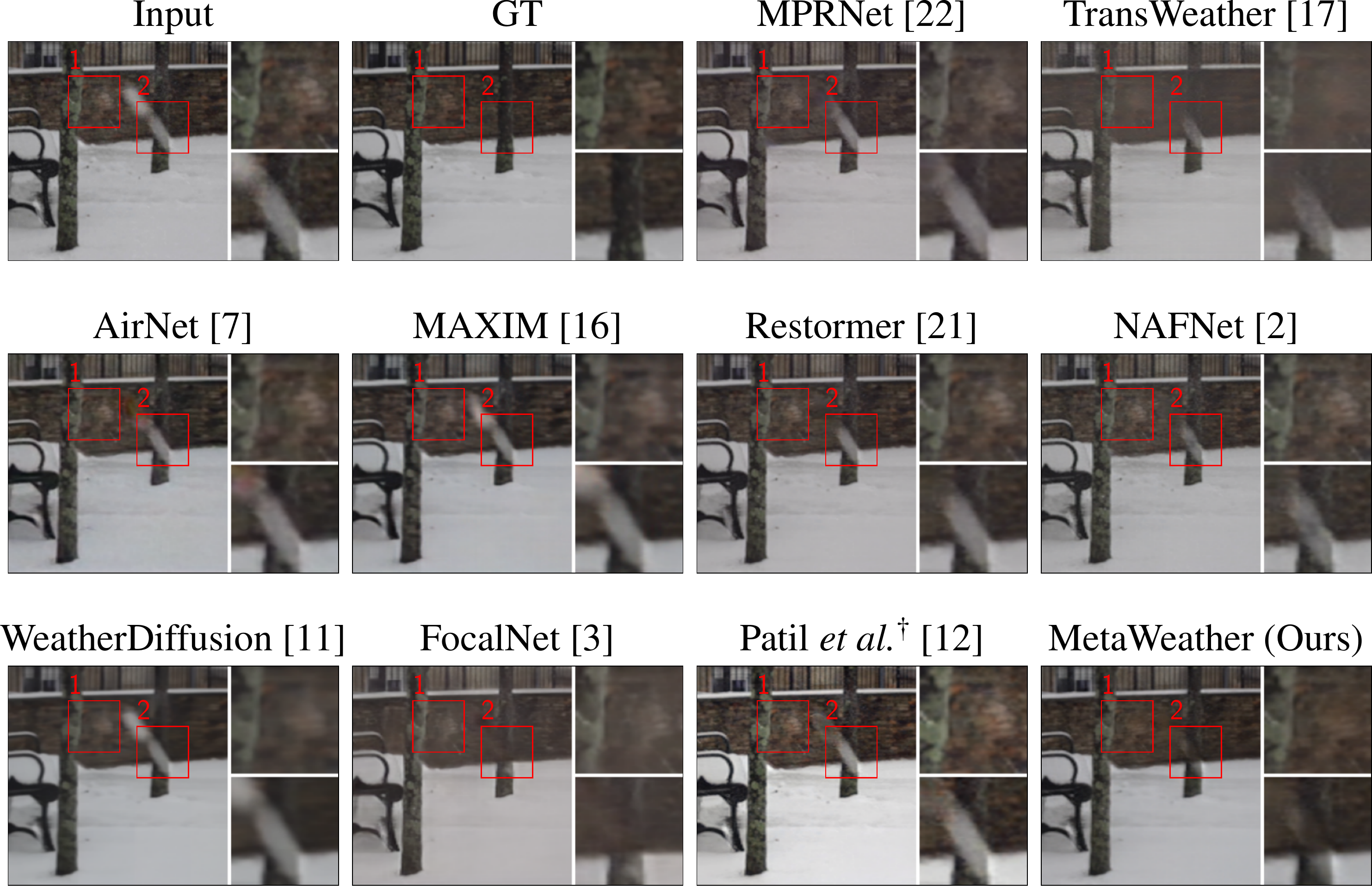}
        \label{fig:suppl-qual-real-snow}
        \caption{RealSnow~\cite{zhu2023learning}}
    \end{subfigure}

    \caption{Qualitative comparison on the SPA-Data~\cite{Wang2019spa} and RealSnow~\cite{zhu2023learning} datasets. $\dagger$ denotes a method for real-world weather-degraded image restoration.
    Best viewed with zoom and color.}
    \label{fig:suppl-qual-real}
\end{figure*}

We provide additional qualitative comparisons to all baselines~\cite{zamir2021multi, zamir2022restormer, tu2022maxim, chen2022simple, cui2023focal, li2022all, valanarasu2022transweather, ozdenizci2023weatherdiffusion, patil2023multi} in \cref{fig:suppl-main-qual} and \cref{fig:suppl-qual-real}.

\section{Further Implementation Details}


\subsection{Details on datasets }


\subsubsection{Meta-train dataset.}
As a meta-train dataset, we use multiple datasets, each containing one of the following degradation patterns: rain, snow, fog, and raindrops.
For the rain dataset, we utilize the Rain13K dataset~\cite{zamir2021multi}, which contains 13,712 clean-rain image pairs. 
This dataset is collected from various existing rainy datasets~\cite{li2016rain,fu2017removing,yang2017deep,zhang2018density,zhang2019image}.
For the snow dataset, we employ the Snow100K dataset, which is proposed in~\cite{Liu2018desnow}.
We use 9,000 pairs in Snow100K, following the setting in~\cite{li2020all,valanarasu2022transweather}.
For the fog dataset, we utilize 10,425 pairs from Foggy Cityscapes~\cite{sakaridis2018semantic}. Lastly, for the raindrop dataset, we used the Raindrop dataset~\cite{qian2018attentive}, which consists of 1,119 pairs. 

\subsubsection{Meta-test and test datasets. }
We use the BID Task~\RNum{2}.A~\cite{han2022blind}, SPA-Data~\cite{Wang2019spa}, and RealSnow~\cite{Zhang2021desnow} datasets for the meta-test phase and the evaluation.
BID Task~\RNum{2}.A dataset~\cite{han2022blind} contains several different combinations of various synthetic weather degradation patterns. 
Each case consists of 500 pairs of degraded images and corresponding ground-truths. We randomly divide each case into two splits: 100 pairs for the meta-test phase and 400 pairs for the evaluation.
We use the test split of the SPA-Data~\cite{Wang2019spa} set, which comprises 1,000 degraded images. We randomly sample 10 pairs for the meta-test phase and 990 pairs for the evaluation. 
Regarding the RealSnow dataset~\cite{Zhang2021desnow}, we use the train split for the meta-test phase, and the test split for the evaluation, following the authors' split. The train and test splits consist of 65,000 and 240 pairs, respectively. We release all splits for each dataset with our code.


\clearpage

\subsection{Further Details}
\subsubsection{Data splitting in the meta-test phase. }
In cases where more than two support set images are given, we interchange their roles between the query image and support set. When only one support pair is available, we utilize it as both the query and support set by duplicating it.

\subsubsection{Training baselines. }
In the pre-training (meta-training) phase, we set the learning rate for the baselines to ${10}^{-4}$. During the adaptation (meta-test) phase, the learning rate is set to ${10}^{-5}$. All models are trained end-to-end in both phases. For the other configurations, we use the same settings as for our method.


\putbib
\end{bibunit}


\begin{thebibliography}{10}
\providecommand{\url}[1]{\texttt{#1}}
\providecommand{\urlprefix}{URL }
\providecommand{\doi}[1]{https://doi.org/#1}

\bibitem{ba2022gtrain}
Ba, Y., Zhang, H., Yang, E., Suzuki, A., Pfahnl, A., Chandrappa, C.C., de~Melo, C.M., You, S., Soatto, S., Wong, A., Kadambi, A.: Not just streaks: Towards ground truth for single image deraining. In: Proceedings of European Conference on Computer Vision. pp. 723--740. Springer (2022)

\bibitem{chen2022simple}
Chen, L., Chu, X., Zhang, X., Sun, J.: Simple baselines for image restoration. In: Proceedings of European Conference on Computer Vision. pp. 17--33. Springer (2022)

\bibitem{cui2023focal}
Cui, Y., Ren, W., Cao, X., Knoll, A.: Focal network for image restoration. In: Proceedings of the IEEE/CVF international conference on computer vision. pp. 13001--13011 (2023)

\bibitem{finn2017model}
Finn, C., Abbeel, P., Levine, S.: Model-agnostic meta-learning for fast adaptation of deep networks. In: Precup, D., Teh, Y.W. (eds.) Proceedings of the 34th International Conference on Machine Learning. Proceedings of Machine Learning Research, vol.~70, pp. 1126--1135. PMLR (06--11 Aug 2017)

\bibitem{gao2021meta}
Gao, X., Wang, Y., Cheng, J., Xu, M., Wang, M.: Meta-learning based relation and representation learning networks for single-image deraining. Pattern Recognition  \textbf{120},  108124 (2021)

\bibitem{gatys2016image}
Gatys, L.A., Ecker, A.S., Bethge, M.: Image style transfer using convolutional neural networks. In: Proceedings of the IEEE conference on computer vision and pattern recognition. pp. 2414--2423 (2016)

\bibitem{han2022blind}
Han, J., Li, W., Fang, P., Sun, C., Hong, J., Armin, M.A., Petersson, L., Li, H.: Blind image decomposition. In: Proceedings of European Conference on Computer Vision. pp. 218--237. Springer (2022)

\bibitem{Hassaballah2021BadWeatherTracking}
Hassaballah, M., Kenk, M.A., Muhammad, K., Minaee, S.: {Vehicle Detection and Tracking in Adverse Weather Using a Deep Learning Framework}. IEEE Transactions on Intelligent Transportation Systems  \textbf{22}(7),  4230--4242 (jul 2021)

\bibitem{Huang2020DSNetWeatherOD}
Huang, S.C., Le, T.H., Jaw, D.W.: {DSNet: Joint Semantic Learning for Object Detection in Inclement Weather Conditions}. IEEE Transactions on Pattern Analysis and Machine Intelligence  \textbf{43}(8),  2623--2633 (2020)

\bibitem{kang2019few}
Kang, B., Liu, Z., Wang, X., Yu, F., Feng, J., Darrell, T.: Few-shot object detection via feature reweighting. In: Proceedings of the IEEE/CVF International Conference on Computer Vision. pp. 8420--8429 (2019)

\bibitem{kim2023universal}
Kim, D., Kim, J., Cho, S., Luo, C., Hong, S.: Universal few-shot learning of dense prediction tasks with visual token matching. In: The Eleventh International Conference on Learning Representations (2023)

\bibitem{lee2022fifo}
Lee, S., Son, T., Kwak, S.: Fifo: Learning fog-invariant features for foggy scene segmentation. In: Proceedings of the IEEE/CVF Conference on Computer Vision and Pattern Recognition. pp. 18911--18921 (2022)

\bibitem{li2022all}
Li, B., Liu, X., Hu, P., Wu, Z., Lv, J., Peng, X.: All-in-one image restoration for unknown corruption. In: Proceedings of the IEEE/CVF Conference on Computer Vision and Pattern Recognition. pp. 17452--17462 (2022)

\bibitem{li2019heavy}
Li, R., Cheong, L.F., Tan, R.T.: Heavy rain image restoration: Integrating physics model and conditional adversarial learning. In: Proceedings of the IEEE/CVF Conference on Computer Vision and Pattern Recognition. pp. 1633--1642 (2019)

\bibitem{li2020all}
Li, R., Tan, R.T., Cheong, L.F.: All in one bad weather removal using architectural search. In: Proceedings of the IEEE/CVF Conference on Computer Vision and Pattern Recognition. pp. 3175--3185 (2020)

\bibitem{Liu2022testtimedehaze}
Liu, H., Wu, Z., Li, L., Salehkalaibar, S., Chen, J., Wang, K.: {Towards Multi-domain Single Image Dehazing via Test-time Training}. In: Proceedings of the IEEE/CVF Conference on Computer Vision and Pattern Recognition. pp. 5821--5830. IEEE (jun 2022)

\bibitem{Liu2022YOLOweather}
Liu, W., Ren, G., Yu, R., Guo, S., Zhu, J., Zhang, L.: {Image-Adaptive YOLO for Object Detection in Adverse Weather Conditions}. Proceedings of the AAAI Conference on Artificial Intelligence  \textbf{36}(2),  1792--1800 (2022)

\bibitem{Liu2018desnow}
Liu, Y.F., Jaw, D.W., Huang, S.C., Hwang, J.N.: {DesnowNet: Context-Aware Deep Network for Snow Removal}. IEEE Transactions on Image Processing  \textbf{27}(6),  3064--3073 (jun 2018)

\bibitem{liu2021swin}
Liu, Z., Lin, Y., Cao, Y., Hu, H., Wei, Y., Zhang, Z., Lin, S., Guo, B.: Swin transformer: Hierarchical vision transformer using shifted windows. In: Proceedings of the IEEE/CVF International Conference on Computer Vision. pp. 10012--10022 (2021)

\bibitem{loshchilov2017decoupled}
Loshchilov, I., Hutter, F.: Decoupled weight decay regularization. arXiv preprint arXiv:1711.05101  (2017)

\bibitem{ozdenizci2023weatherdiffusion}
{\"O}zdenizci, O., Legenstein, R.: Restoring vision in adverse weather conditions with patch-based denoising diffusion models. IEEE Transactions on Pattern Analysis and Machine Intelligence  (2023)

\bibitem{patil2023multi}
Patil, P.W., Gupta, S., Rana, S., Venkatesh, S., Murala, S.: Multi-weather image restoration via domain translation. In: Proceedings of the IEEE/CVF International Conference on Computer Vision. pp. 21696--21705 (2023)

\bibitem{qian2018attentive}
Qian, R., Tan, R.T., Yang, W., Su, J., Liu, J.: Attentive generative adversarial network for raindrop removal from a single image. In: Proceedings of the IEEE/CVF Conference on Computer Vision and Pattern Recognition. pp. 2482--2491 (2018)

\bibitem{Qian2018raindrop}
Qian, R., Tan, R.T., Yang, W., Su, J., Liu, J.: {Attentive Generative Adversarial Network for Raindrop Removal from A Single Image}. In: Proceedings of the IEEE/CVF Conference on Computer Vision and Pattern Recognition. pp. 2482--2491. IEEE (jun 2018)

\bibitem{rai2022fluid}
Rai, S.N., Saluja, R., Arora, C., Balasubramanian, V.N., Subramanian, A., Jawahar, C.: Fluid: Few-shot self-supervised image deraining. In: Proceedings of the IEEE/CVF Winter Conference on Applications of Computer Vision. pp. 3077--3086 (2022)

\bibitem{ran2023few}
Ran, W., Yuan, W., Shibasaki, R.: Few-shot depth completion using denoising diffusion probabilistic model. In: Proceedings of the IEEE/CVF Conference on Computer Vision and Pattern Recognition. pp. 6558--6566 (2023)

\bibitem{ronneberger2015u}
Ronneberger, O., Fischer, P., Brox, T.: U-net: Convolutional networks for biomedical image segmentation. In: Proceedings of Medical Image Computing and Computer-Assisted Intervention--MICCAI 2015: 18th International Conference. pp. 234--241. Springer (2015)

\bibitem{sakaridis2018semantic}
Sakaridis, C., Dai, D., Van~Gool, L.: Semantic foggy scene understanding with synthetic data. International Journal of Computer Vision  \textbf{126},  973--992 (2018)

\bibitem{shaban2017one}
Shaban, A., Bansal, S., Liu, Z., Essa, I., Boots, B.: One-shot learning for semantic segmentation. In: Proceedings of the British Machine Vision Conference. pp. 167.1--167.13. BMVA Press (September 2017)

\bibitem{Tremblay2021BadWeatherSegmentDepth}
Tremblay, M., Halder, S.S., de~Charette, R., Lalonde, J.F.: {Rain Rendering for Evaluating and Improving Robustness to Bad Weather}. International Journal of Computer Vision  \textbf{129}(2),  341--360 (feb 2021)

\bibitem{tu2022maxim}
Tu, Z., Talebi, H., Zhang, H., Yang, F., Milanfar, P., Bovik, A., Li, Y.: Maxim: Multi-axis mlp for image processing. In: Proceedings of the IEEE/CVF Conference on Computer Vision and Pattern Recognition. pp. 5769--5780 (2022)

\bibitem{ulyanov2016instance}
Ulyanov, D., Vedaldi, A., Lempitsky, V.: Instance normalization: The missing ingredient for fast stylization. arXiv preprint arXiv:1607.08022  (2016)

\bibitem{valanarasu2022transweather}
Valanarasu, J.M.J., Yasarla, R., Patel, V.M.: Transweather: Transformer-based restoration of images degraded by adverse weather conditions. In: Proceedings of the IEEE/CVF Conference on Computer Vision and Pattern Recognition. pp. 2353--2363 (2022)

\bibitem{vinyals2016matching}
Vinyals, O., Blundell, C., Lillicrap, T., Wierstra, D., et~al.: Matching networks for one shot learning. Proceedings of Advances in Neural Information Processing Systems  \textbf{29} (2016)

\bibitem{wang2019panet}
Wang, K., Liew, J.H., Zou, Y., Zhou, D., Feng, J.: Panet: Few-shot image semantic segmentation with prototype alignment. In: Proceedings of the IEEE/CVF International Conference on Computer Vision. pp. 9197--9206 (2019)

\bibitem{Wang2019spa}
Wang, T., Yang, X., Xu, K., Chen, S., Zhang, Q., Lau, R.W.: {Spatial Attentive Single-Image Deraining With a High Quality Real Rain Dataset}. In: Proceedings of the IEEE/CVF Conference on Computer Vision and Pattern Recognition. pp. 12262--12271. IEEE (jun 2019)

\bibitem{wang2022uformer}
Wang, Z., Cun, X., Bao, J., Zhou, W., Liu, J., Li, H.: Uformer: A general u-shaped transformer for image restoration. In: Proceedings of the IEEE/CVF Conference on Computer Vision and Pattern Recognition. pp. 17683--17693 (2022)

\bibitem{xie2022simmim}
Xie, Z., Zhang, Z., Cao, Y., Lin, Y., Bao, J., Yao, Z., Dai, Q., Hu, H.: Simmim: A simple framework for masked image modeling. In: Proceedings of the IEEE/CVF Conference on Computer Vision and Pattern Recognition. pp. 9653--9663 (2022)

\bibitem{ye2023adverse}
Ye, T., Chen, S., Bai, J., Shi, J., Xue, C., Jiang, J., Yin, J., Chen, E., Liu, Y.: Adverse weather removal with codebook priors. In: Proceedings of the IEEE/CVF International Conference on Computer Vision. pp. 12653--12664 (2023)

\bibitem{zamir2022restormer}
Zamir, S.W., Arora, A., Khan, S., Hayat, M., Khan, F.S., Yang, M.H.: Restormer: Efficient transformer for high-resolution image restoration. In: Proceedings of the IEEE/CVF Conference on Computer Vision and Pattern Recognition. pp. 5728--5739 (2022)

\bibitem{zamir2020learning}
Zamir, S.W., Arora, A., Khan, S., Hayat, M., Khan, F.S., Yang, M.H., Shao, L.: Learning enriched features for real image restoration and enhancement. In: Computer Vision--ECCV 2020: 16th European Conference, Glasgow, UK, August 23--28, 2020, Proceedings, Part XXV 16. pp. 492--511. Springer (2020)

\bibitem{zamir2021multi}
Zamir, S.W., Arora, A., Khan, S., Hayat, M., Khan, F.S., Yang, M.H., Shao, L.: Multi-stage progressive image restoration. In: Proceedings of the IEEE/CVF Conference on Computer Vision and Pattern Recognition. pp. 14821--14831 (2021)

\bibitem{zhao2023comprehensive}
Zhao, H., Gou, Y., Li, B., Peng, D., Lv, J., Peng, X.: Comprehensive and delicate: An efficient transformer for image restoration. In: Proceedings of the IEEE/CVF Conference on Computer Vision and Pattern Recognition. pp. 14122--14132 (2023)

\bibitem{zhou2023fourmer}
Zhou, M., Huang, J., Guo, C.L., Li, C.: Fourmer: An efficient global modeling paradigm for image restoration. In: International Conference on Machine Learning. pp. 42589--42601. PMLR (2023)

\bibitem{zhu2023learning}
Zhu, Y., Wang, T., Fu, X., Yang, X., Guo, X., Dai, J., Qiao, Y., Hu, X.: Learning weather-general and weather-specific features for image restoration under multiple adverse weather conditions. In: Proceedings of the IEEE/CVF Conference on Computer Vision and Pattern Recognition. pp. 21747--21758 (2023)

\end{thebibliography}


\begin{thebibliography}{10}
\providecommand{\url}[1]{\texttt{#1}}
\providecommand{\urlprefix}{URL }
\providecommand{\doi}[1]{https://doi.org/#1}

\bibitem{zaken2021bitfit}
Ben~Zaken, E., Goldberg, Y., Ravfogel, S.: {B}it{F}it: Simple parameter-efficient fine-tuning for transformer-based masked language-models. In: Proceedings of the 60th Annual Meeting of the Association for Computational Linguistics (Volume 2: Short Papers). pp.~1--9. Association for Computational Linguistics, Dublin, Ireland (May 2022)

\bibitem{chen2022simple}
Chen, L., Chu, X., Zhang, X., Sun, J.: Simple baselines for image restoration. In: Proceedings of European Conference on Computer Vision. pp. 17--33. Springer (2022)

\bibitem{cui2023focal}
Cui, Y., Ren, W., Cao, X., Knoll, A.: Focal network for image restoration. In: Proceedings of the IEEE/CVF international conference on computer vision. pp. 13001--13011 (2023)

\bibitem{fu2017removing}
Fu, X., Huang, J., Zeng, D., Huang, Y., Ding, X., Paisley, J.: Removing rain from single images via a deep detail network. In: Proceedings of the IEEE/CVF Conference on Computer Vision and Pattern Recognition. pp. 3855--3863 (2017)

\bibitem{han2022blind}
Han, J., Li, W., Fang, P., Sun, C., Hong, J., Armin, M.A., Petersson, L., Li, H.: Blind image decomposition. In: Proceedings of European Conference on Computer Vision. pp. 218--237. Springer (2022)

\bibitem{kim2023universal}
Kim, D., Kim, J., Cho, S., Luo, C., Hong, S.: Universal few-shot learning of dense prediction tasks with visual token matching. In: The Eleventh International Conference on Learning Representations (2023)

\bibitem{li2022all}
Li, B., Liu, X., Hu, P., Wu, Z., Lv, J., Peng, X.: All-in-one image restoration for unknown corruption. In: Proceedings of the IEEE/CVF Conference on Computer Vision and Pattern Recognition. pp. 17452--17462 (2022)

\bibitem{li2020all}
Li, R., Tan, R.T., Cheong, L.F.: All in one bad weather removal using architectural search. In: Proceedings of the IEEE/CVF Conference on Computer Vision and Pattern Recognition. pp. 3175--3185 (2020)

\bibitem{li2016rain}
Li, Y., Tan, R.T., Guo, X., Lu, J., Brown, M.S.: Rain streak removal using layer priors. In: Proceedings of the IEEE/CVF Conference on Computer Vision and Pattern Recognition. pp. 2736--2744 (2016)

\bibitem{Liu2018desnow}
Liu, Y.F., Jaw, D.W., Huang, S.C., Hwang, J.N.: {DesnowNet: Context-Aware Deep Network for Snow Removal}. IEEE Transactions on Image Processing  \textbf{27}(6),  3064--3073 (jun 2018)

\bibitem{ozdenizci2023weatherdiffusion}
{\"O}zdenizci, O., Legenstein, R.: Restoring vision in adverse weather conditions with patch-based denoising diffusion models. IEEE Transactions on Pattern Analysis and Machine Intelligence  (2023)

\bibitem{patil2023multi}
Patil, P.W., Gupta, S., Rana, S., Venkatesh, S., Murala, S.: Multi-weather image restoration via domain translation. In: Proceedings of the IEEE/CVF International Conference on Computer Vision. pp. 21696--21705 (2023)

\bibitem{qi2022parameter}
Qi, W., Ruan, Y.P., Zuo, Y., Li, T.: Parameter-efficient tuning on layer normalization for pre-trained language models. arXiv preprint arXiv:2211.08682  (2022)

\bibitem{qian2018attentive}
Qian, R., Tan, R.T., Yang, W., Su, J., Liu, J.: Attentive generative adversarial network for raindrop removal from a single image. In: Proceedings of the IEEE/CVF Conference on Computer Vision and Pattern Recognition. pp. 2482--2491 (2018)

\bibitem{sakaridis2018semantic}
Sakaridis, C., Dai, D., Van~Gool, L.: Semantic foggy scene understanding with synthetic data. International Journal of Computer Vision  \textbf{126},  973--992 (2018)

\bibitem{tu2022maxim}
Tu, Z., Talebi, H., Zhang, H., Yang, F., Milanfar, P., Bovik, A., Li, Y.: Maxim: Multi-axis mlp for image processing. In: Proceedings of the IEEE/CVF Conference on Computer Vision and Pattern Recognition. pp. 5769--5780 (2022)

\bibitem{valanarasu2022transweather}
Valanarasu, J.M.J., Yasarla, R., Patel, V.M.: Transweather: Transformer-based restoration of images degraded by adverse weather conditions. In: Proceedings of the IEEE/CVF Conference on Computer Vision and Pattern Recognition. pp. 2353--2363 (2022)

\bibitem{wang2019panet}
Wang, K., Liew, J.H., Zou, Y., Zhou, D., Feng, J.: Panet: Few-shot image semantic segmentation with prototype alignment. In: Proceedings of the IEEE/CVF International Conference on Computer Vision. pp. 9197--9206 (2019)

\bibitem{Wang2019spa}
Wang, T., Yang, X., Xu, K., Chen, S., Zhang, Q., Lau, R.W.: {Spatial Attentive Single-Image Deraining With a High Quality Real Rain Dataset}. In: Proceedings of the IEEE/CVF Conference on Computer Vision and Pattern Recognition. pp. 12262--12271. IEEE (jun 2019)

\bibitem{yang2017deep}
Yang, W., Tan, R.T., Feng, J., Liu, J., Guo, Z., Yan, S.: Deep joint rain detection and removal from a single image. In: Proceedings of the IEEE/CVF Conference on Computer Vision and Pattern Recognition. pp. 1357--1366 (2017)

\bibitem{zamir2022restormer}
Zamir, S.W., Arora, A., Khan, S., Hayat, M., Khan, F.S., Yang, M.H.: Restormer: Efficient transformer for high-resolution image restoration. In: Proceedings of the IEEE/CVF Conference on Computer Vision and Pattern Recognition. pp. 5728--5739 (2022)

\bibitem{zamir2021multi}
Zamir, S.W., Arora, A., Khan, S., Hayat, M., Khan, F.S., Yang, M.H., Shao, L.: Multi-stage progressive image restoration. In: Proceedings of the IEEE/CVF Conference on Computer Vision and Pattern Recognition. pp. 14821--14831 (2021)

\bibitem{zhang2018density}
Zhang, H., Patel, V.M.: Density-aware single image de-raining using a multi-stream dense network. In: Proceedings of the IEEE/CVF Conference on Computer Vision and Pattern Recognition. pp. 695--704 (2018)

\bibitem{zhang2019image}
Zhang, H., Sindagi, V., Patel, V.M.: Image de-raining using a conditional generative adversarial network. IEEE transactions on circuits and systems for video technology  \textbf{30}(11),  3943--3956 (2019)

\bibitem{Zhang2021desnow}
Zhang, K., Li, R., Yu, Y., Luo, W., Li, C.: {Deep Dense Multi-Scale Network for Snow Removal Using Semantic and Depth Priors}. IEEE Transactions on Image Processing  \textbf{30},  7419--7431 (2021)

\bibitem{zhu2023learning}
Zhu, Y., Wang, T., Fu, X., Yang, X., Guo, X., Dai, J., Qiao, Y., Hu, X.: Learning weather-general and weather-specific features for image restoration under multiple adverse weather conditions. In: Proceedings of the IEEE/CVF Conference on Computer Vision and Pattern Recognition. pp. 21747--21758 (2023)

\end{thebibliography}
\end{document}